\documentclass[runningheads]{llncs}

\usepackage{eccv}

\usepackage{eccvabbrv}
\usepackage{tcolorbox}
\usepackage{listings}
\usepackage{multirow}

\usepackage{graphicx}
\usepackage{booktabs}
\usepackage{xspace}
\everypar\expandafter{\the\everypar\looseness=-1}
\newcommand{\appcref}[1]{Appendix~\ref{#1}}

\usepackage{ccicons}

\usepackage[accsupp]{axessibility}  

\usepackage{hyperref}

\usepackage{tikz}
\usepackage{pgfplots}
\usepackage{pgfplotstable}
\pgfplotsset{compat=1.16}
\usetikzlibrary{patterns}

\definecolor{colArtSAGENet}{HTML}{1f77b4}
\definecolor{colCLIP}{HTML}{9467bd}
\definecolor{colColPali}{HTML}{2ca02c}
\definecolor{colColQwen}{HTML}{17becf}
\definecolor{colMSC}{HTML}{ff7f0e}
\definecolor{colSigLIP}{HTML}{d62728}
\definecolor{colCANVAS}{HTML}{e377c2}

\usepackage{orcidlink}
\usepackage{xspace}

\usepackage{enumitem}
\usepackage{xr-hyper}
\newcommand{\approach}{\textbf{CANVAS}\xspace}

\newcommand{\huggingface}{\raisebox{-1.5pt}{\includegraphics[height=1.05em]{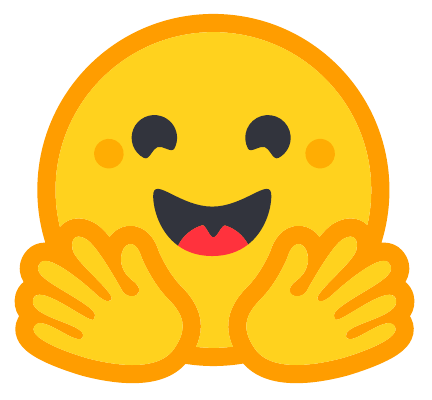}}\xspace}
\newcommand{\github}{\raisebox{-1.5pt}{\includegraphics[height=1.05em]{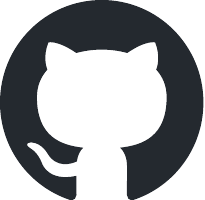}}\xspace}

\begin{document}

\title{Art Beyond Semantics: Sheaf-Informed Contrastive Learning for Multi-Relational Representations} 

\titlerunning{Sheaf-Informed Contrastive Learning for Multi-Relational Representations}

\author{Ludovica Schaerf\inst{1,2}$^{,*}$\orcidlink{0000-0001-9460-702X} \and
Antonio Purificato\inst{3,4}$^{,*}$$^{,\dagger}$\orcidlink{0009-0009-3933-380X} \and Piera Riccio\inst{5}\orcidlink{0000-0001-8602-8271} \and \\
Fabrizio Silvestri\inst{3}\orcidlink{0000-0001-7669-9055} \and Noa Garcia\inst{6}\orcidlink{0000-0002-9200-6359}}

\authorrunning{L. Schaerf et al.}

\institute{University of Zurich, Switzerland \and
Max Planck Institute Bibliotheca Hertziana, Italy \and
Sapienza University of Rome, Italy \and 
Amazon, Luxembourg \and
University of Amsterdam, Netherlands \and
The University of Osaka, Japan\\
\email{ludovica.schaerf@uzh.ch} \\
$*$ Equal contribution $\dagger$ Work done outside of Amazon \\
{\small\huggingface{}\href{https://huggingface.co/datasets/antoniopuri/SemArtPlus}{\textsc{SemArt+}}} \quad
{\small\huggingface{}\href{https://huggingface.co/datasets/antoniopuri/WikiArtPlus}{\textsc{WikiArt+}}} \quad
{\small\github{}\hspace{0.5em}\href{https://github.com/antoniopurificato/artistic_sheaf/tree/stable}{\textsc{Code}}}}

\maketitle
\setcounter{footnote}{0}
\renewcommand{\thefootnote}{\alph{footnote}}

\begin{abstract}
Understanding a painting is never a single act. Art historians may analyze the same work through concepts of style, iconography, or historical context, dimensions that are not interchangeable, and each carries distinct semantic relationships between the visual and the textual. Vision-Language Models (VLMs) like CLIP, which learn a single shared embedding space, collapse this richness into a single homogeneous alignment, thereby losing the multi-relational structure that defines art-historical reasoning. 
We introduce \approach (\textbf{C}ontrastive \textbf{A}rt-aware \textbf{N}etwork for \textbf{V}ision-Language \textbf{A}lignment with \textbf{S}heaves), a framework for learning relation-aware multimodal representations inspired by sheaf theory. Each artwork is projected into multiple embeddings conditioned on the type of relation (\ie, the context), and a novel contrastive loss encodes contextual information during training, with no dependency on external data at inference. We evaluate on three newly introduced benchmarks of artworks for multi-relational art understanding: \textsc{WikiArt+}, derived from WikiArt and Wikipedia, \textsc{HertzianaDP}, from the Bibliotheca Hertziana collection, and \textsc{SemArt+}, refined from the \textsc{SemArt} dataset. In multimodal retrieval and art understanding, \approach outperforms the baselines, supporting the view that multi-relational alignment is not just theoretically motivated but also practically essential.
\end{abstract}

\section{Introduction}

Before computer vision had massive collections of images, it had art. Early classification algorithms were tested on digitized paintings, aiming to distinguish a master's brushstroke from a copyist's \cite{doi:10.1073/pnas.0406398101}. With the rise of large-scale online datasets, the field has since shifted focus to photorealistic images \cite{deng2009imagenet}, which are easier to obtain and less open to interpretation. However, art remains a distinct challenge by itself \cite{castellano2021deep}: while standard image analysis might focus on accurate object detection  \cite{zou2023object}, the art domain is concerned with decoding the nuanced language of style, intent, meaning, and context that defines each artwork \cite{castellano2021deep}.

\begin{figure}[t]
    \centering
    \includegraphics[width=0.95\linewidth]{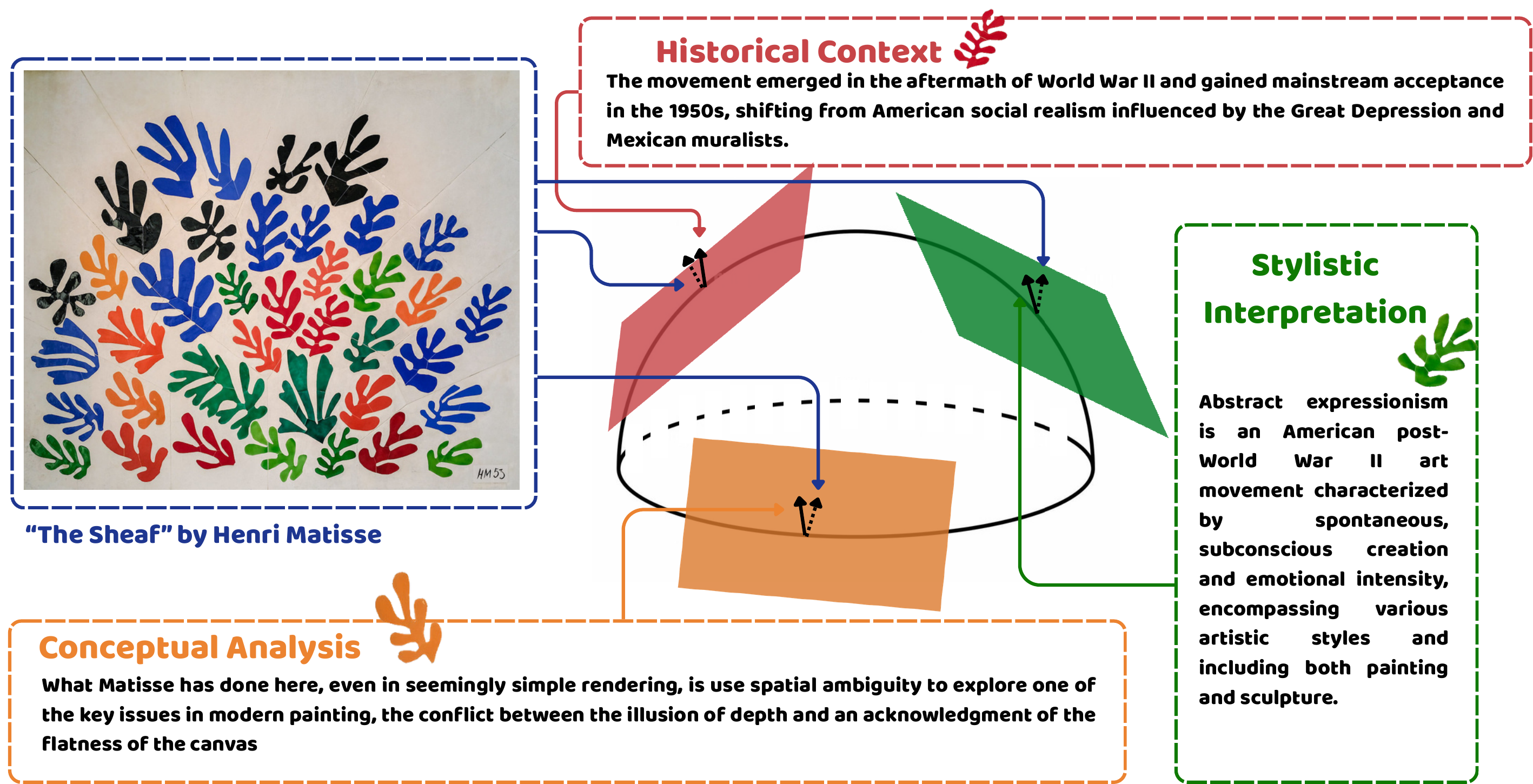}
    \caption{Matisse’s painting ``La Gerbe'' (``The Sheaf'') from \textsc{WikiArt+} dataset alongside three modes of art-historical understanding: Historical Context (\eg, post-war artistic shifts), Stylistic Interpretation (\eg, traits of the broader movement), and Conceptual Analysis (\eg, the artist's use of spatial ambiguity). We show how \approach encodes the images and corresponding texts into subspaces corresponding to the different modes of interpretation. Artwork in the public domain. \ccPublicDomain}
    \label{fig:aspects}
\end{figure}

Current vision-language models (VLMs) \cite{clip} trained via contrastive learning fundamentally assume a single fixed relationship between images and their associated texts \cite{NEURIPS2022_2cd36d32}. This limits the capabilities of these models in domains in which multiple valid interpretations per image can coexist. Art history is a canonical example: a single artwork can simultaneously engage with different types of descriptions. As shown in \cref{fig:aspects}, Matisse's painting ``La Gerbe'' (``The Sheaf'') can be, for instance, associated with texts referring to historical context (\eg, situating the work within post-war European modernism and the broader shift of the avant-garde toward the United States), stylistic interpretation (\eg, referring to recurrent elements characterizing the movement this painting belongs to), or conceptual analysis (\eg, interpreting the meaning of the painting), each constituting a distinct yet interrelated mode of understanding the artwork itself \cite{castellano2021deep}. %

In an attempt to capture the different interpretations of art in context, graph-based approaches~\cite {efthymiou2021graph,scaringi2025graphclip} propose encoding images across multiple aspects and modeling relational structures for classification and regression. Fundamentally, these approaches require access to the full graph structure at test time, including edge indices derived from test-set annotations \cite{hamilton2017inductive}. This means that they operate in a \textit{transductive} setting, where predictions are only possible for images already integrated into the graph, and unseen images cannot be evaluated unless they are first annotated and added. This setting constrains their applicability in \textit{inductive} scenarios, where previously unseen entities must be processed without known relational connections, a common requirement in real-world settings~\cite {teru2020inductive}.

To overcome the limited capacity for multi-semanticity in VLMs and the constraints of graph-based approaches, we propose \textbf{C}ontrastive \textbf{A}rt-aware \textbf{N}etwork for \textbf{V}ision-Language \textbf{A}lignment with \textbf{S}heaves (\approach), a method that combines VLM representations with Sheaf graph formulations \cite{bodnar2022neural}. \approach encodes images and text using a pre-trained CLIP \cite{clip} model and projects them into relation-aware subspaces using graph-based distances. This allows the network to learn multiple representations within a single space without requiring the graph connectivity at test time. \approach works as a lightweight fine-tuning. Rather than representing each image and text as a single representation, it models them as a set, each capturing a distinct dimension of the link between an artwork and an associated text.

We validate \approach for cross-modal retrieval and classification accuracy on three art historical datasets: (1) the \textsc{SemArt+} dataset \cite{garcia2018read} consisting of European paintings from the 3rd to 19th centuries, augmented with per-sentence annotations from \textsc{Explain Me the Painting} \cite{bai2021explain}; (2) the \textsc{WikiArt+} dataset, an extension of \textsc{WikiArt} \cite{artgan2018} consisting of worldwide art, enriched with Wikipedia-sourced explanations; and (3) a newly introduced dataset, 
\textsc{HertzianaDP}\cite{3.Z8W2JR_2025}, consisting of a collection of paintings (P) and drawings (D) belonging to the Bibliotheca Hertziana, not publicly available prior to this work.\footnote{\url{https://edmond.mpg.de/dataset.xhtml?persistentId=doi:10.17617/3.1GN3OL}\newline \url{https://edmond.mpg.de/dataset.xhtml?persistentId=doi:10.17617/3.Z8W2JR}} 
As, to the best of our knowledge, several large VLMs, such as OpenCLIP, have been exposed to the first two datasets during training \cite{ramos2025data}, \textsc{HertzianaDP} serves as an ideal testbed for evaluating generalization to unseen artworks. Across all datasets, \approach consistently outperforms baseline VLMs and existing adaptation methods. In image-to-text retrieval, \approach outperforms the strongest baseline by a large margin across all datasets, with the biggest gains on \textsc{HertzianaDP}, which was not online at the time of CLIP pretraining. \approach also delivers strong results in text-to-image retrieval and relation-aware classification, with no other baseline matching its consistency.

\section{Related work}
\label{sec:related}
This work combines multimodal representation learning with graph-based modeling in the context of art understanding. In the following, we first introduce how computer vision has been applied to art, and then we review relevant work on vision–language contrastive learning and graph-augmented representations.

\subsection{Computer vision for art understanding}
Early large-scale studies~\cite{karayev2013recognizing, crowley2014state, van2017learning} demonstrate that deep convolutional networks trained on natural images can be effectively transferred to artistic domains, with a focus on a single domain-specific task. For example, Karayev \etal~\cite{karayev2013recognizing} explore style prediction across photographs and paintings, while Crowley \etal~\cite{crowley2014state} investigate object retrieval and representation learning in artworks. Similarly, van Noord \etal~\cite{van2017learning} study artist attribution with deep features.
Subsequent work~\cite{saleh2015large,strezoski2017omniart} instead builds on larger, more comprehensive benchmarks. Saleh \etal~\cite{saleh2015large} analyze together style, genre, and artist classification on thousands of paintings, highlighting specific challenges of artistic imagery. Later, Strezoski \etal~\cite{strezoski2017omniart}, introduce the OmniArt dataset, unifying multiple art-related prediction tasks and encouraging shared visual representations across them. 


\subsection{Vision-language contrastive learning}
Vision-language contrastive models, such as CLIP \cite{clip}, align images and texts in a joint embedding space. However, they incur several problems in the context of our work: (1) CLIP's contrastive loss treats all non-paired samples as negatives, pushing apart semantically related but non-matched image-text pairs within each batch, and (2) it enforces single global alignments instead of capturing multiple valid relationships between image regions or across images.
Several works~\cite{10.1145/3627673.3679619 , zhu2022relclip, pan2022contrastive, qiao2025multimodalrepresentationlearningconditioned} address the latter by refining the contrastive objective itself. Xie \etal~\cite{10.1145/3627673.3679619} introduce a Main Semantics Consistency loss to prioritize learning the most semantically relevant aspect of an image or text, while Zhu \etal~\cite{zhu2022relclip} and Pan \etal~\cite{pan2022contrastive} move beyond single-semantic, exploring relation-level visual-semantic alignment within images through relational contrastive learning. Qiao \etal~\cite{qiao2025multimodalrepresentationlearningconditioned}, extends the former by introducing a relation-guided cross-attention mechanism that modulates multimodal representations between images under each relation context, using natural-language relation descriptions, while Ruthardt\etal~\cite{ruthardt2026steerable} use cross-attention to steer the CLIP representation toward the desired textual cue.A different extension of CLIP to multiple representations of the same image through text can be found in document retrieval models, such as ColPali and ColQwen~\cite{fayssecolpali}. These vision-language models are trained to produce multi-vector embeddings from document page images, directly optimizing the downstream retrieval objective via a late-interaction mechanism. These works allow for a coherent representation of textual additions within images. 
Despite their effectiveness, these contrastive approaches only represent within-image relational content and purely semantic relations. In the context of art, we wish to model multiple relations, including non-semantic ones.

\subsection{Graph-augmented representations}
Graph-augmented representation learning injects contextual knowledge, such as stylistic or historical connections, into learned embeddings, enriching them beyond simple image–text pairs. Some works have leveraged these possibilities in the context of art.
A common strategy employs graph neural networks (GNNs) to propagate information across entity neighborhoods and enrich per-node representations. ArtSAGENet~\cite{efthymiou2021graph} and ContextNet~\cite{garcia2019context,garcia2020contextnet} combine visual features with GNN representations that operate over artist-level relationships, thereby inserting contextual information into the embeddings. Furthermore, El Vaigh \etal~\cite{el2025gnnboost} adopt a transductive GNN approach on a knowledge graph of artworks to jointly predict multiple metadata labels for unlabelled samples. Lastly, Scaringi \etal~\cite{scaringi2025graphclip} replaces CLIP’s text encoder with a GNN over a knowledge graph and learns a shared image-graph embedding space. While these graph-augmented methods leverage contextual information, they do not address multi-relational multimodal alignment. Some of the works~\cite{efthymiou2021graph, el2025gnnboost, scaringi2025graphclip}, furthermore, require the full graph to be available at inference time, preventing their application to unseen entities.


\section{Contrastive Art-aware Network for Vision-Language Alignment with Sheaves (\approach)}

Our approach is based on an intuition: since an artwork and a text can be linked by different relationships that depend on both the visual and textual content, the way they are compared should adapt accordingly. For example, consider a painting paired with two different texts, one describing its \textit{visual content} (``a portrait of a woman in Renaissance dress'') and another describing its \textit{historical context} (``commissioned by the Medici family in 1490''). We wish to learn distinct transformations for each relation: the content relationship should align the image embedding with visual-semantic features in the text space, while the context relationship should project the same image into a historical-factual subspace. Rather than centering the embeddings of each item, our model treats relationships as ``first-class citizens'': what matters most is not the item representations per se, but the maps that transport them into relation-specific subspaces.

This intuition can be naturally formalized as a graph through cellular sheaf theory \cite{barbero2022sheaf}. In a cellular sheaf, nodes and edges of the graph are equipped with vector spaces (stalks), and restriction maps transform features between nodes' coordinate systems that pass through the edge, enabling the same data to be consistently represented in different local frames.  
Unlike standard GNNs, sheaf-inspired restriction maps provide a mechanism for learning how to transport both image and text representations into different edge-based subspaces, effectively deforming the embedding space to align with each relation. 

\subsection{Preliminaries}
We first review cellular sheaf theory's key concepts. In a cellular sheaf, data is assigned to both the nodes and edges of a graph, enabling the joint modeling of local and global structure. 
Given an undirected graph $G = (V, E)$ \cite{bondy2008graph}, a cellular sheaf $\mathcal{F}$ associates to each node $v \in V$ a vector space $\mathcal{F}(v)$, and to each edge $e \in E$ a vector space $\mathcal{F}(e)$. These spaces represent the local data supported on nodes and edges, respectively.

For every incident node-edge pair $v \unlhd e$, the sheaf defines a linear restriction map $\mathcal{F}_{v \unlhd e}: \mathcal{F}(v) \rightarrow \mathcal{F}(e)$, which encodes how information attached to a node is related to the information attached to the edge. 
The vector spaces $\mathcal{F}(v)$ and $\mathcal{F}(e)$ are referred to as \textit{stalks}. Restriction maps are fundamental, as they determine how local data is consistently transported across the graph and thus provide the mechanism that connects local representations to global structure.

\subsection{Method}
\begin{figure}[t]
    \centering
    \includegraphics[width=0.8\linewidth]{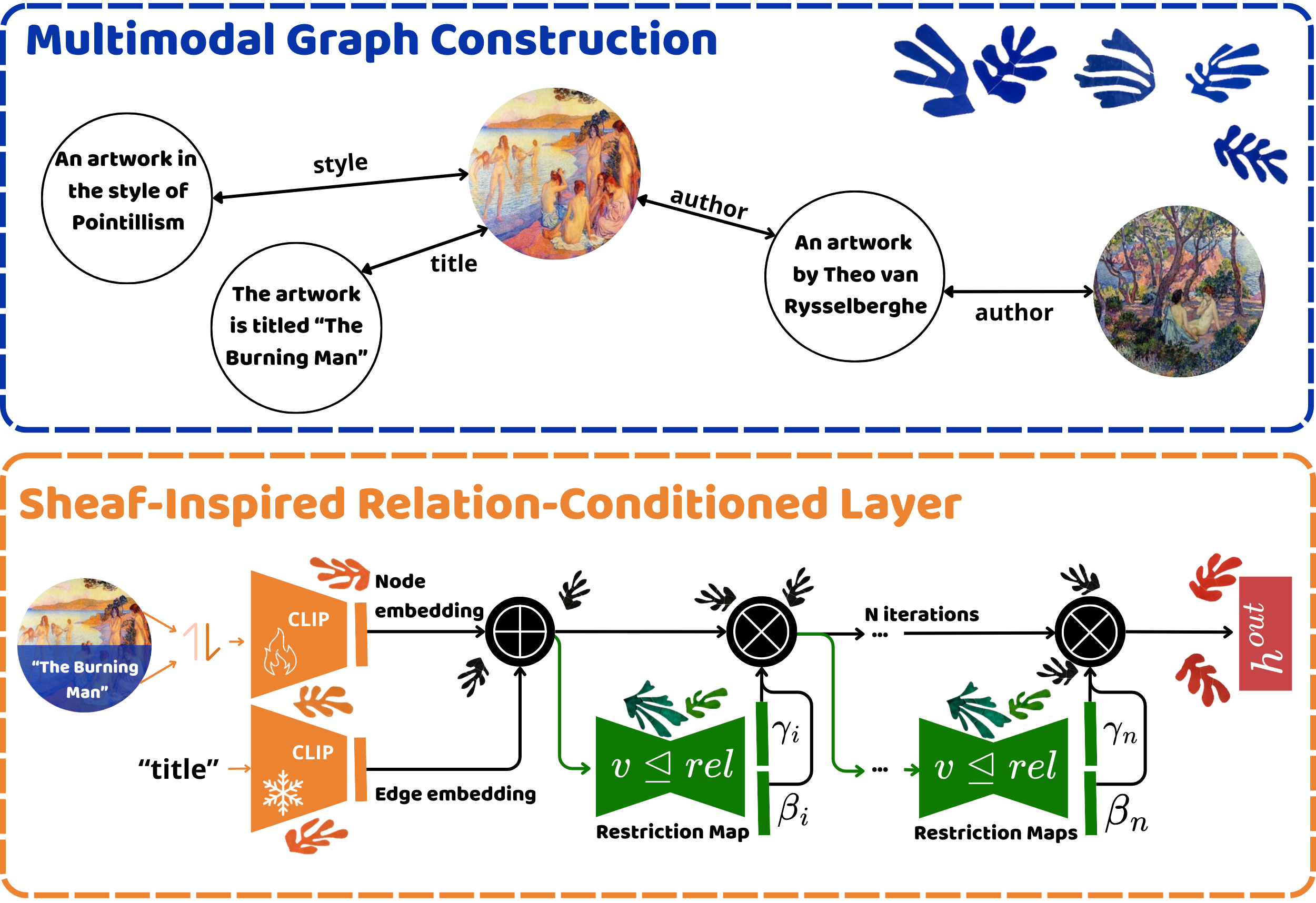}
    \caption{\textbf{Overview of \approach}. Top: The input data is structured as a bipartite graph in which image nodes (artworks) and text nodes (descriptions) are connected via typed edges representing semantic relationships (\eg, ``style'', ``author''). Bottom: An image or text in a node is encoded using CLIP, alongside its relationship, the edge type. The node embedding is concatenated with the relation embedding and passed through learned restriction maps. These maps output relation-specific scale ($\gamma_{i}$) and shift ($\beta_{i}$) parameters that transform the node representations into relation-aware subspaces via affine modulation. This process is repeated N times. The $\otimes$ and $\oplus$ symbols denote the element-wise multiplication and concatenation operations, respectively.}
    \label{graphical_abstract_part1}
\end{figure}

We adopt \textit{restriction maps} as learnable edge-conditioned transformations that project node representations into relation-specific edge subspaces, allowing a single image or text embedding to be compared differently depending on the relationship it participates in. Additionally, we introduce a context-aware soft contrastive loss derived from the graph structure. This objective enforces instance-level alignment between image–text pairs within a given relation type, while softening the negative term based on graph proximity: nodes that are nearby in the graph receive a weaker penalty than fully unconnected ones, reflecting the intuition that items sharing more relationships are semantically closer. We present the input graph formation and Sheaf-inspired relation-conditioned layer in \cref{graphical_abstract_part1}, while the loss is schematized in \cref{graphical_abstract_part2}. 

\subsection{Multimodal graph construction}
As shown in \cref{graphical_abstract_part1}, rather than treating each image–text pair in isolation, we model the dataset as a bipartite graph where edges encode typed semantic relations. This graph determines which pairs interact during training and provides the relational context that conditions the learned embeddings.

We model multimodal data as a bipartite graph $G = (U \cup V, E)$,  where $U$ denotes image nodes, $V$ denotes text nodes, and $E \subseteq U \times V$ encodes semantic relations between images and texts.
Each edge $e = (u,v) \in E$ is associated with a relation label $r_e$ (\eg, \emph{content}, \emph{context}), represented as the name of the relation.

\paragraph{CLIP initialization.}
We initialize node and edge features using a pretrained CLIP model~\cite{clip}. 
Given an image $u \in U$, a text $v \in V$, and a relation description $r_e$ for edge $e \in E$, we compute:
\begin{equation}
\mathbf{h}_u = f_{\text{img}}(u), \quad
\mathbf{h}_v = f_{\text{text}}(v), \quad
\mathbf{h}_{r_e} = f_{\text{text}}(r_e),
\end{equation}
where $f_{\text{img}}$ and $f_{\text{text}}$ denote the CLIP image and text encoders.

After initialization, we fine-tune the projection layers and the last transformer blocks, while keeping earlier layers frozen. After the projection layer within the CLIP model, the $\mathbf{h}_u$ and $\mathbf{h}_v$ are embedded into a shared latent space of size $d$.

\subsection{Sheaf-inspired relation-conditioned layer}
Nodes connected by different relation types should be compared in different representational subspaces. Inspired by cellular sheaf theory, we learn a relation-conditioned transformation that modulates each node embedding depending on the relation, producing distinct relation-aware views without separate encoders. We illustrate this mechanism in \cref{graphical_abstract_part1}.

Given an edge $e = (u,v)$ with relation embedding $\mathbf{h}_{r_e}$, we want to produce a transformation that depends on both the node content and the relation type. We implement this process as a feature-wise linear modulation (FiLM) \cite{perez2018film}: we first concatenate each node embedding (either the image embedding $\mathbf{h}_u$ or the text embedding $\mathbf{h}_v$) with the relation embedding $\mathbf{h}_{r_e}$.
\begin{equation}
\mathbf{z}_{u,e} = [\mathbf{h}_u \oplus \mathbf{h}_{r_e}], 
\quad
\mathbf{z}_{v,e} = [\mathbf{h}_v \oplus \mathbf{h}_{r_e}],
\end{equation}
where $\oplus$ denotes concatenation. We define a sheaf learner $\phi: \mathbb{R}^{d + d_r} \rightarrow \mathbb{R}^{2d}$
implemented as a three-layer MLP with LeakyReLU activations \cite{bodnar2022neural}. 
The output is split into feature-wise scale and shift parameters:
\begin{equation}
(\boldsymbol{\gamma}_e, \boldsymbol{\beta}_e)
=
\phi(\mathbf{z}_{\cdot,e}),
\quad
\boldsymbol{\gamma}_e, \boldsymbol{\beta}_e \in \mathbb{R}^d.
\end{equation}

The restriction map is applied as a FiLM modulation:
\begin{equation}
\mathbf{\tilde{h}}_{e, u}
=
\boldsymbol{\gamma}_e \otimes \mathbf{h_u}
+
\boldsymbol{\beta}_e,
\quad
\mathbf{\tilde{h}}_{e, v}
=
\boldsymbol{\gamma}_e \otimes \mathbf{h_v}
+
\boldsymbol{\beta}_e,
\end{equation}
where $\otimes$ denotes element-wise multiplication.
Intuitively, $\boldsymbol{\gamma}_e$ amplifies or suppresses dimensions of the embedding that are relevant to the relation, while $\boldsymbol{\beta}_e$ shifts the representation to align with the relation-specific subspace. The same node embedding is transformed differently for each relation, yielding a learned restriction map that locally deforms the embedding space for each relation type.

We stack $N$ layers and average the outputs, as common practice in GNNs~\cite{he2020lightgcn}:
\begin{equation}
\mathbf{\tilde{h}}_{e,u}^{\text{out}}
=
\frac{1}{N}
\sum_{n=1}^{N}
\mathbf{\tilde{h}}_{e,u}^{(n)},
\quad
\mathbf{\tilde{h}}_{e,v}^{\text{out}}
=
\frac{1}{N}
\sum_{n=1}^{N}
\mathbf{\tilde{h}}_{e,v}^{(n)}.
\end{equation}

Unlike standard GNN message passing, where each node aggregates information from all its neighbors, our formulation processes each image–text edge independently: the restriction map transforms a node embedding conditioned solely on the relation associated with that edge. This ensures that embeddings can be computed for individual image–text pairs at inference time, without access to the full graph structure.

\subsection{Graph-based soft similarity}
\begin{figure}[t]
    \centering
    \includegraphics[width=\linewidth]{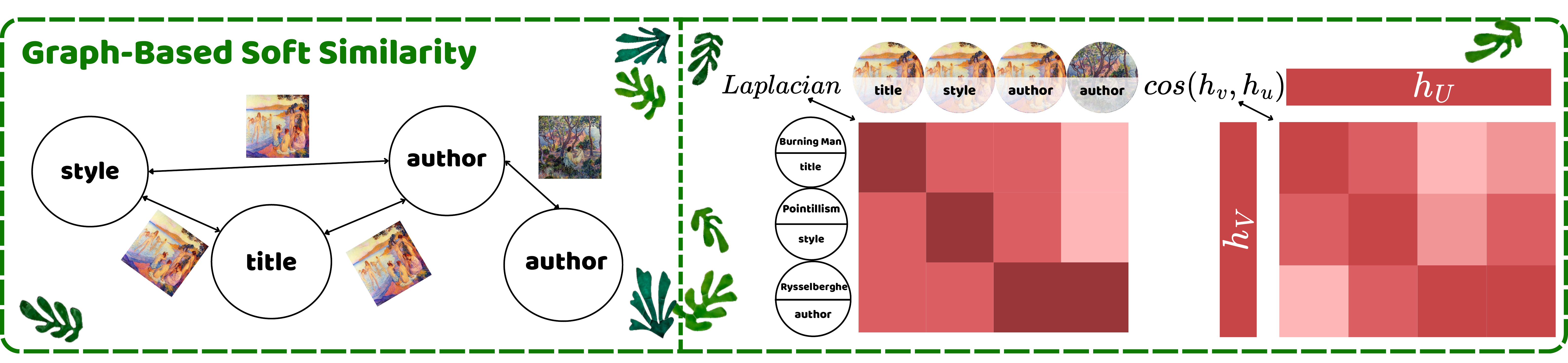}
    \caption{\textbf{Graph-based soft similarity computation via line graph construction and heat kernel diffusion.} Left panel: Illustration of the line graph transformation $G'=L(G)$, where the relationships associated with the edges become the nodes and vice versa. Right panel: The heatmap on the left visualizes the soft target matrix computed via heat kernel diffusion on the line graph's Laplacian. Pairs involving the same image or text (\eg, ``Burning Man-title'', ``Burning Man-style'') receive higher scores, reflecting their semantic proximity in the graph structure. On the right, the pairwise cosine similarities between image-relation and text-relation embeddings outputted by the sheaf layer in \cref{graphical_abstract_part1}. The contrastive loss is computed between the soft target matrix and the cosine similarity matrix. Darker red indicates higher similarity.}
    \label{graphical_abstract_part2}
\end{figure}
The InfoNCE contrastive objective aligns image–text pairs independently, without considering how different relationship types relate to one another. To capture higher-order dependencies between relations, we turn to the structure of the graph itself.
We construct the line graph $G' = \mathcal{L}(G) = (V', E')$~\cite{hoffman1964line}, in which each node corresponds to an edge in the original graph $V' = E$, and two nodes are connected whenever their corresponding edges $E'=\{e_i,e_j\}\subseteq E$ share a common endpoint $e_i\cap e_j\neq\emptyset$. Intuitively, the line graph makes the relationships between relationships explicit: two image–text pairs become neighbors in $G'$ if they involve the same image or the same text. We illustrate this construction in \cref{graphical_abstract_part2}.
We then use the line graph to derive a soft structural prior that captures how closely related any two image–text pairs are. Let $L_{G'} = D' - A'$ be the combinatorial Laplacian of $G'$, with degree matrix $D'$ and adjacency matrix $A'$~\cite{bondy2008graph}. We compute contextual affinities via the heat kernel:
\begin{equation}
W = \exp(-\tau L_{G'}),
\end{equation}
where $\tau > 0$ controls the extent of diffusion: small values preserve only immediate neighbors, while larger values propagate affinity across more distant pairs. To obtain a sharp, normalized distribution over neighbors, we raise each entry to a power $\alpha$ and row-normalize:
\begin{equation}
\tilde{W}{ij} = \frac{(W{ij})^\alpha}{\sum_k (W_{ik})^\alpha}.
\end{equation}

The resulting matrix $\tilde{W}$ serves as a soft target that modulates the contrastive loss: pairs that are structurally close in $G'$ are penalized less strongly as negatives, reflecting the intuition that image–text pairs sharing common items are semantically closer.

\subsection{Relation-aware contrastive learning}
Our training objective combines two signals: a contrastive term that aligns matching image-relation-text pairs and a soft term that encourages the learned similarities to be consistent with the relational topology of the line graph.

Let $\mathbf{H}_U$ and $\mathbf{H}_V$ be the final image and text matrices, where $\mathbf{H}_U$ contains all the output representations $\mathbf{\tilde{h}}_{e,u}^{\mathrm{out}}$ of the (relation type, image) pairs, while $\mathbf{H}_V$ contains all $\mathbf{\tilde{h}}_{e,v}^{\mathrm{out}}$ of the (relation type, text) pairs.

We $\ell_2$-normalize embeddings before computing similarities:
\begin{equation}
S_{UV} = \mathbf{H}_U \mathbf{H}_V^\top,
\quad
S_{VU} = \mathbf{H}_V \mathbf{H}_U^\top.
\end{equation}

\paragraph{CLIP Loss.}

We adopt the symmetric InfoNCE objective~\cite{oord2018representation}:
\begin{equation}
\mathcal{L}_{\text{CLIP}}
=
\frac{1}{2}
\left(
\mathcal{L}_{\text{InfoNCE}}(S_{UV})
+
\mathcal{L}_{\text{InfoNCE}}(S_{VU})
\right).
\end{equation}

\paragraph{Graph-regularized KL loss.}

We construct a soft contextual target:
\begin{equation}
P=\lambda \tilde{W}+(1-\lambda) I,
\end{equation}
where $\lambda \in [0,1]$ is a mixing coefficient that controls the trade-off between graph-based contextual similarity and strict identity matching, and $I$ is the identity matrix. Following \emph{Gao et al.}~\cite{gao2024softclip}, we minimize the symmetric KL divergence to soften the CLIP loss:
\begin{equation}
\mathcal{L}_{\text{KL}}
=
\frac{1}{2}
\left(
\mathrm{KL}(P \,\|\, S_{UV})
+
\mathrm{KL}(P^\top \,\|\, S_{VU})
\right).
\end{equation}

\paragraph{Final objective.}

The overall loss balances contrastive alignment and contextual relations:
\begin{equation}
\mathcal{L}
=
(1 - \eta)\mathcal{L}_{\text{CLIP}}
+
\eta \mathcal{L}_{\text{KL}},
\end{equation}
where $\eta \in [0,1]$ balances the contrastive objective and the graph-regularized divergence. Overall, \approach combines relation-conditioned restriction maps with a context-informed contrastive objective. This unifies local relational transformations and global regularization for relation-aware multimodal alignment.
Intuitively, the loss encourages the representations of each image-relation pair to align with the corresponding text-relation pair while addressing CLIP's hard-negatives problem when negative pairs are not uncorrelated (i.e., either different relations of the same item or items that are connected). 

\section{Experiments}

\paragraph{Baselines}
We compare \approach against approaches that incorporate semantic or graph-structured information for multimodal retrieval and art understanding:
\begin{itemize}[label=$\bullet$]
    \item ArtSAGENet~\cite{efthymiou2021graph}: it combines visual feature learning with a knowledge-enhanced graph component to improve multi-task visual representation learning in art. At test time, we remove the graph-connectivity dependency, as explained in \cite{garcia2020contextnet}, for a fair comparison.
    \item CLIP \cite{clip} and SigLIP \cite{siglip}: they learn joint image-text embeddings using a contrastive objective, enabling zero-shot cross-modal retrieval and classification. SigLIP extends CLIP using a sigmoid-based loss. Throughout the paper, pretrained models are denoted as CLIP and SigLIP, whereas fine-tuned variants are denoted as CLIP-ft and SigLIP-ft.
    \item ColPali and ColQwen~\cite{fayssecolpali}: these models produce multi-vector embeddings from document page images and rely on late-interaction matching mechanisms to directly optimize downstream retrieval tasks.
     \item GraphCLIP~\cite{scaringi2025graphclip}: extends CLIP by aligning artwork images with knowledge-graph representations, enabling context-aware artwork classification.
    \item MSC~\cite{10.1145/3627673.3679619}: it introduces a semantically optimized retrieval framework based on a Main Semantics Consistency loss. Their objective is to rank the semantically most relevant cross-modal matches during retrieval.
    \item RCML~\cite{qiao2025multimodalrepresentationlearningconditioned}: learns image–text embeddings by conditioning contrastive learning on semantic relations using relation-guided cross-attention.
\end{itemize}

\paragraph{Datasets}
We report results on 3 datasets for multi-relational art understanding:
\begin{itemize}[label=$\bullet$]
    \item \textsc{HertzianaDP}: consisting of a collection of paintings and drawings belonging to the Bibliotheca HertzianaDP, not publicly available prior to this work, which was processed to obtain a mix of textual and metadata-based fields \cite{3.Z8W2JR_2025}; it consists of 45,819 images and 86,321 texts split into 125,267 nodes and 190,195 edges.
    \item \textsc{SemArt+}~\cite{garcia2018read}: we process the \textsc{SemArt} dataset augmented with per-sentence annotations from \textsc{Explain Me the Paintings} \cite{bai2021explain} through metadata cleaning and pronoun resolution (as documented in \cref{sec:pronounres}); it contains 34,770 images and 62,289 texts split in 97,053 nodes and 151,430 edges.
    \item \textsc{WikiArt+}:  we extend \textsc{WikiArt}~\cite{artgan2018} by enriching artworks, artists, and art movements with processed Wikipedia-sourced explanations; it consists of 22,028 images and 29,603 texts split into 51,631 nodes and 308,050 edges.
\end{itemize}

The data is split into train, validation, and test sets, ensuring each item appears only once to prevent leakage. Additional details are in \appcref{app:exp_setup}.
  
\paragraph{Implementation details}
All experiments are implemented in PyTorch. As CLIP implementation, we use OpenCLIP ViT-B/32 pre-trained on LAION-2B \cite{clip}. We apply a partial fine-tuning strategy: the last $L_{\text{ft}}=3$ transformer blocks of the vision and text encoders, along with the projection heads, are unfrozen; all remaining parameters are kept frozen. Edge-relation embeddings are produced with the fully frozen text encoder. Computational cost is detailed in \appcref{app:exp_setup}.

The core of \approach consists of $N$ stacked sheaf-inspired layers. Each layer learns restriction maps via a three-layer MLP
Sheaf Learner, with hidden dimensions $512$ and $256$ and LeakyReLU activations. We set the latent dimension to $d = 512$ and stack $N = 3$ layers. We use the AdamW optimizer with a learning rate of $10^{-5}$ and a batch size of $256$. We train for $50$ epochs; we apply gradient clipping by norm with a maximum value of~$1.0$ and employ early stopping with a patience of~$5$ epochs, monitoring the validation loss. We report results on the test set using the model that achieved the best results on the validation set. The parameters we select are: $\tau = 0.7$, $\alpha = 1.2$, $\lambda = 0.7$, and $\eta = 0.5$. At inference time, \approach only requires the image or text and, optionally, a relation type. The relation type can be inferred from text as reported in \cref{tab:link_pred_accuracy}.

\paragraph{Evaluation}
Following prior work \cite{efthymiou2021graph,10.1145/3627673.3679619}, we compute Recall@$K$ (fraction of queries with a correct match in the top-$K$), Precision@$K$ (fraction of relevant items in the top-$K$), and NDCG@$K$ (which gives higher credit to correct matches appearing earlier in the
ranked list) for retrieval, and Accuracy for classification.
For Recall, Precision, and NDCG, we evaluate image-to-text (I2T) and text-to-image (T2I) scores. Due to space constraints, only Recall results are shown; additional results are in \appcref{app:additional_results}.

\section{Results}

\begin{table*}[t]
\centering
\caption{Retrieval results in terms of Recall (R) in text-to-image (T2I) and image-to-text (I2T) across the selected datasets ($K=\{5,10\}$). \textbf{Bold} denotes the best model for a dataset, \underline{underlined} the second best.}
\label{tab:main_results_retrieval}
\resizebox{\linewidth}{!}{
\begin{tabular}{lcccc|cccc|cccc}
\toprule
& \multicolumn{4}{c}{\textsc{HertzianaDP}} & \multicolumn{4}{c}{\textsc{SemArt+}} & \multicolumn{4}{c}{\textsc{WikiArt+}} \\
\cmidrule(lr){2-5}  \cmidrule(lr){6-9} \cmidrule(lr){10-13}
 Method & T2I R@5 & T2I R@10 & I2T R@5 & I2T R@10 & T2I R@5 & T2I R@10 & I2T R@5 & I2T R@10 & T2I R@5 & T2I R@10 & I2T R@5 & I2T R@10 \\
\midrule
ArtSAGENet & 0.001 & 0.001 & 0.000 & 0.001 & 0.003 & 0.004 & 0.002 & 0.003 & 0.001 & 0.002 & 0.000 & 0.000 \\
CLIP & 0.041 & 0.063 & 0.013 & 0.018 & 0.305 & 0.385 & 0.078 & 0.111 & 0.131 & 0.174 & 0.042 & 0.059 \\
CLIP-ft & 0.046 & 0.069 & \underline{0.057} & \underline{0.084} & 0.189 & 0.216 & 0.093 & 0.131 & 0.137 & 0.149 & 0.082 & 0.158 \\
ColPali & 0.011 & 0.017 & 0.003 & 0.006 & 0.123 & 0.177 & 0.021 & 0.028 & 0.043 & 0.061 & 0.008 & 0.011 \\
ColQwen & 0.007 & 0.011 & 0.001 & 0.002 & 0.154 & 0.206 & 0.025 & 0.037 & 0.037 & 0.054 & 0.008 & 0.011 \\
MSC & 0.000 & 0.001 & 0.000 & 0.000 & 0.006 & 0.010 & 0.012 & 0.023 & 0.002 & 0.003 & 0.011 & 0.024 \\
RCML & 0.026 & 0.041 & 0.044 & 0.071 & 0.134 & 0.163 & 0.076 & 0.105 & 0.017 & 0.025 & 0.080 & 0.148 \\
SigLIP & \underline{0.061} & \underline{0.083} & 0.016 & 0.022 & 0.334 & 0.423 & 0.095 & 0.133 & \textbf{0.239} & 0.301 & 0.067 & 0.091 \\
SigLIP-ft & 0.038 & 0.058 & 0.051 & 0.076 & \underline{0.337} & \underline{0.446} & \underline{0.102} & \underline{0.150} & 0.137 & \textbf{0.307} & \underline{0.115} & \underline{0.209} \\
\approach & \textbf{0.099} & \textbf{0.158} & \textbf{0.386} & \textbf{0.514} & \textbf{0.376} & \textbf{0.489} & \textbf{0.645} & \textbf{0.714} & \underline{0.144} & \underline{0.217} & \textbf{0.703} & \textbf{0.781} \\
\bottomrule
\end{tabular}
}
\end{table*}

\def\dataHertzianasagenet{(0,0.0009) (1,0.0000) (2,0.0009) (3,0.0006)}
\def\dataHertzianaclip{(0,0.0070) (1,0.1452) (2,0.1973) (3,0.0098)}
\def\dataHertzianaclipft{(0,0.0074) (1,0.1533) (2,0.2318) (3,0.0117)}
\def\dataHertzianacolpali{(0,0.0048) (1,0.1093) (2,0.0380) (3,0.0064)}
\def\dataHertzianacolqwen2{(0,0.0059) (1,0.0789) (2,0.0190) (3,0.0062)}
\def\dataHertzianamsc{(0,0.0044) (1,0.0054) (2,0.0051) (3,0.0030)}
\def\dataHertzianarcml{(0,0.0734) (1,0.1692) (2,0.2082) (3,0.0544)}
\def\dataHertzianasiglip{(0,0.0068) (1,0.2226) (2,0.2368) (3,0.0056)}
\def\dataHertzianasiglipft{(0,0.0079) (1,0.2234) (2,0.2833) (3,0.0089)}
\def\dataHertzianasheafclip{(0,0.1132) (1,0.1242) (2,0.3010) (3,0.0980)}
\def\dataSemArtsagenet{(0,0.0000) (1,0.0016) (2,0.0010) (3,0.0000)}
\def\dataSemArtclip{(0,0.4724) (1,0.4287) (2,0.5547) (3,0.5797)}
\def\dataSemArtclipft{(0,0.4813) (1,0.4435) (2,0.5831) (3,0.5431)}
\def\dataSemArtcolpali{(0,0.3182) (1,0.2048) (2,0.3066) (3,0.3949)}
\def\dataSemArtcolqwen2{(0,0.3225) (1,0.2104) (2,0.3639) (3,0.4130)}
\def\dataSemArtmsc{(0,0.0146) (1,0.0132) (2,0.0146) (3,0.0616)}
\def\dataSemArtrcml{(0,0.3794) (1,0.4419) (2,0.4902) (3,0.5206)}
\def\dataSemArtsiglip{(0,0.4921) (1,0.4871) (2,0.4932) (3,0.5072)}
\def\dataSemArtsiglipft{(0,0.4634) (1,0.4713) (2,0.5412) (3,0.5753)}
\def\dataSemArtsheafclip{(0,0.4434) (1,0.4857) (2,0.5881) (3,0.6123)}
\def\dataWikidatasetsagenet{(0,0.0661) (1,0.0411) (2,0.1937) (3,0.1834)}
\def\dataWikidatasetclip{(0,0.2407) (1,0.2850) (2,0.0527) (3,0.0391)}
\def\dataWikidatasetclipft{(0,0.2913) (1,0.3133) (2,0.0611) (3,0.0321)}
\def\dataWikidatasetcolpali{(0,0.0975) (1,0.1088) (2,0.0149) (3,0.0159)}
\def\dataWikidatasetcolqwen2{(0,0.1325) (1,0.1040) (2,0.0182) (3,0.0176)}
\def\dataWikidatasetmsc{(0,0.0326) (1,0.0433) (2,0.0034) (3,0.0035)}
\def\dataWikidatasetrcml{(0,0.3774) (1,0.4067) (2,0.1141) (3,0.1283)}
\def\dataWikidatasetsiglip{(0,0.3900) (1,0.4347) (2,0.0713) (3,0.0485)}
\def\dataWikidatasetsiglipft{(0,0.4419) (1,0.4926) (2,0.1256) (3,0.1611)}
\def\dataWikidatasetsheafclip{(0,0.5315) (1,0.5535) (2,0.1795) (3,0.1757)}

\definecolor{colArtSAGENet}{HTML}{1f77b4}
\definecolor{colCLIP}{HTML}{9467bd}
\definecolor{colColPali}{HTML}{2ca02c}
\definecolor{colColQwen}{HTML}{17becf}
\definecolor{colMSC}{HTML}{ff7f0e}
\definecolor{colSigLIP}{HTML}{d62728}
\definecolor{colCANVAS}{HTML}{e377c2}
\definecolor{colRCML}{HTML}{8c564b}
\definecolor{colCLIPFT}{HTML}{D4AF37}      
\definecolor{colSigLIPFT}{HTML}{4D4D4D}    

\pgfplotsset{
    compat=1.11,
    every tick label/.append style={font=\small},
    ybar legend/.style={
        legend image code/.code={
            \fill[#1] (3pt,-3pt) rectangle (5pt,3pt);
        }
    }
}

\pgfplotsset{
    recallbar/.style={
        ybar=0pt,
        bar width=2pt,
        width=0.41\textwidth,
        height=0.28\textwidth,
        ymin=0,
        ymax=0.7,
        enlarge x limits=0.2,
        ymajorgrids=true,
        major grid style={solid, gray, opacity=0.2},
        tick label style={font=\scriptsize},
        label style={font=\small},
        title style={font=\small, yshift=-6pt},
    }
}

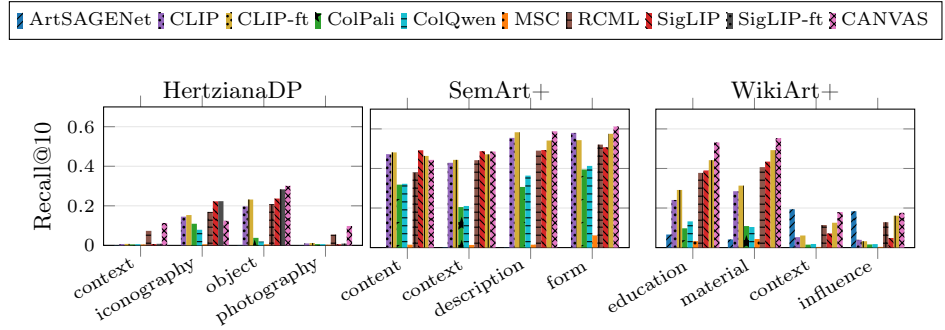
\begin{figure}[h]
\centering

\begin{tikzpicture}[baseline=(current bounding box.north)]
\begin{axis}[
    hide axis,
    scale only axis,
    height=0pt,
    width=0pt,
    legend style={
        draw=black,
        cells={anchor=west},
        legend columns=10,
        column sep=0.03em,
        font=\scriptsize,
        at={(0.5,1)},
        anchor=south,
    },
    legend entries={
        ArtSAGENet, CLIP, CLIP-ft, ColPali, ColQwen, MSC, RCML, SigLIP, SigLIP-ft, CANVAS
    },
]
    \addlegendimage{ybar, fill=colArtSAGENet, draw=none, postaction={pattern=north east lines}, ybar legend}
    \addlegendimage{ybar, fill=colCLIP, draw=none, postaction={pattern=crosshatch dots}, ybar legend}
    \addlegendimage{ybar, fill=colCLIPFT, draw=none, postaction={pattern=crosshatch dots}, ybar legend}
    \addlegendimage{ybar, fill=colColPali, draw=none, postaction={pattern=fivepointed stars}, ybar legend}
    \addlegendimage{ybar, fill=colColQwen, draw=none, postaction={pattern=horizontal lines}, ybar legend}
    \addlegendimage{ybar, fill=colMSC, draw=none, postaction={pattern=dots}, ybar legend}
    \addlegendimage{ybar, fill=colRCML, draw=none, postaction={pattern=grid}, ybar legend}
    \addlegendimage{ybar, fill=colSigLIP, draw=none, postaction={pattern=north west lines}, ybar legend}
     \addlegendimage{ybar, fill=colSigLIPFT, draw=none, postaction={pattern=crosshatch dots}, ybar legend}
    \addlegendimage{ybar, fill=colCANVAS, draw=none, postaction={pattern=crosshatch}, ybar legend}
    \addplot[draw=none] coordinates {(0,0)};
\end{axis}
\end{tikzpicture}

\vspace{4pt}

\setlength{\tabcolsep}{0.05pt}
\begin{tabular}{ccc}

\begin{tikzpicture}[baseline=(current bounding box.north)]
\begin{axis}[
    recallbar,
    trim axis right,
    title={HertzianaDP},
    ylabel={Recall@10},
    ytick={0,0.2,0.4,0.6,0.8,1},
    xtick={0,1,2,3},
    xticklabels={context, iconography, object, photography},
    xticklabel style={rotate=30, anchor=east},
]
    \addplot[fill=colArtSAGENet, draw=none, postaction={pattern=north east lines}] coordinates {\dataHertzianasagenet};
    \addplot[fill=colCLIP, draw=none, postaction={pattern=crosshatch dots}] coordinates {\dataHertzianaclip};
    \addplot[fill=colCLIPFT, draw=none,
    postaction={pattern=vertical lines}]
    coordinates {\dataHertzianaclipft};
    \addplot[fill=colColPali, draw=none, postaction={pattern=fivepointed stars}] coordinates {\dataHertzianacolpali};
    \addplot[fill=colColQwen, draw=none, postaction={pattern=horizontal lines}] coordinates {\dataHertzianacolqwen2};
    \addplot[fill=colMSC, draw=none, postaction={pattern=dots}] coordinates {\dataHertzianamsc};
    \addplot[fill=colRCML, draw=none, postaction={pattern=grid}] coordinates {\dataHertzianarcml};
    \addplot[fill=colSigLIP, draw=none, postaction={pattern=north west lines}] coordinates {\dataHertzianasiglip};
    \addplot[fill=colSigLIPFT, draw=none,
    postaction={pattern=vertical lines}]
    coordinates {\dataHertzianasiglipft};
    \addplot[fill=colCANVAS, draw=none, postaction={pattern=crosshatch}] coordinates {\dataHertzianasheafclip};
\end{axis}
\end{tikzpicture}

\hspace{-7mm}

\begin{tikzpicture}[baseline=(current bounding box.north)]
\begin{axis}[
    recallbar,
    trim axis right,
    trim axis left,
    title={SemArt+},
    yticklabels=\empty,
    xtick={0,1,2,3},
    xticklabels={content, context, description, form},
    xticklabel style={rotate=30, anchor=east},
]
    \addplot[fill=colArtSAGENet, draw=none, postaction={pattern=north east lines}] coordinates {\dataSemArtsagenet};
    \addplot[fill=colCLIP, draw=none, postaction={pattern=crosshatch dots}] coordinates {\dataSemArtclip};
    \addplot[fill=colCLIPFT, draw=none,
    postaction={pattern=vertical lines}]
    coordinates {\dataSemArtclipft};
    \addplot[fill=colColPali, draw=none, postaction={pattern=fivepointed stars}] coordinates {\dataSemArtcolpali};
    \addplot[fill=colColQwen, draw=none, postaction={pattern=horizontal lines}] coordinates {\dataSemArtcolqwen2};
    \addplot[fill=colMSC, draw=none, postaction={pattern=dots}] coordinates {\dataSemArtmsc};
    \addplot[fill=colRCML, draw=none, postaction={pattern=grid}] coordinates {\dataSemArtrcml};
    \addplot[fill=colSigLIP, draw=none, postaction={pattern=north west lines}] coordinates {\dataSemArtsiglip};
    \addplot[fill=colCLIPFT, draw=none,
    postaction={pattern=vertical lines}]
    coordinates {\dataSemArtsiglipft};
    \addplot[fill=colCANVAS, draw=none, postaction={pattern=crosshatch}] coordinates {\dataSemArtsheafclip};
\end{axis}
\end{tikzpicture}

\hspace{-7mm}

\begin{tikzpicture}[baseline=(current bounding box.north)]
\begin{axis}[
    recallbar,
    trim axis right,
    title={WikiArt+},
    yticklabels=\empty,
    xtick={0,1,2,3},
    xticklabels={education, material, context, influence},
    xticklabel style={rotate=30, anchor=east},
]
    \addplot[fill=colArtSAGENet, draw=none, postaction={pattern=north east lines}] coordinates {\dataWikidatasetsagenet};
    \addplot[fill=colCLIP, draw=none, postaction={pattern=crosshatch dots}] coordinates {\dataWikidatasetclip};
    \addplot[fill=colCLIPFT, draw=none,
    postaction={pattern=vertical lines}]
    coordinates {\dataWikidatasetclipft};
    \addplot[fill=colColPali, draw=none, postaction={pattern=fivepointed stars}] coordinates {\dataWikidatasetcolpali};
    \addplot[fill=colColQwen, draw=none, postaction={pattern=horizontal lines}] coordinates {\dataWikidatasetcolqwen2};
    \addplot[fill=colMSC, draw=none, postaction={pattern=dots}] coordinates {\dataWikidatasetmsc};
    \addplot[fill=colRCML, draw=none, postaction={pattern=grid}] coordinates {\dataWikidatasetrcml};
    \addplot[fill=colSigLIP, draw=none, postaction={pattern=north west lines}] coordinates {\dataWikidatasetsiglip};
    \addplot[fill=colCLIPFT, draw=none,
    postaction={pattern=vertical lines}]
    coordinates {\dataWikidatasetsiglipft};
    \addplot[fill=colCANVAS, draw=none, postaction={pattern=crosshatch}] coordinates {\dataWikidatasetsheafclip};
\end{axis}
\end{tikzpicture}

\end{tabular}
\caption{Text-to-Image Recall@10 by relation type across the three datasets.}
\label{fig:retrieval_split}
\end{figure}

\begin{figure}[t]
    \centering
    \includegraphics[width=0.9\linewidth]{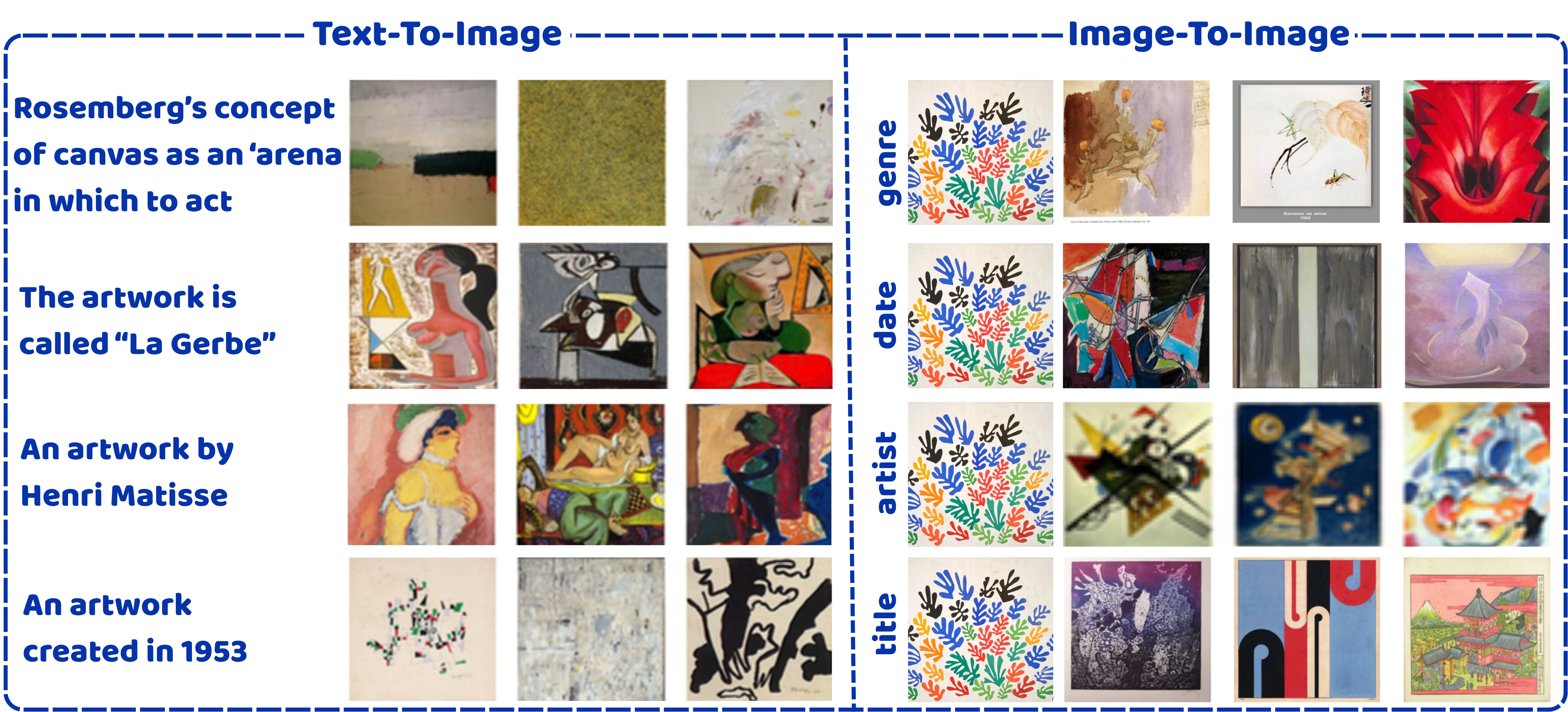}
    \caption{\textbf{Qualitative examples of relation-aware multimodal retrieval using \approach.} Left panel (text-to-image): Each text query activates a different relation-specific subspace, yielding diverse retrieval results: Rosenberg's observation obtains gestural and materic artworks, the title is associated with an early cubist tradition, while artist queries return other Matisse works that are stylistically similar to his Fauvist period, and date queries return contemporaneous artworks from the early 50's. Right panel (image-to-image): Given ``The Sheaf'' as a visual query, \approach retrieves relevant images across different relation types: genre finds decorative works featuring plants, date retrieves artworks from the 40s and 50s, artist finds abstract expressionist works with similar compositions, and title recovers thematically related pieces.}
    \label{fig:qualitative_examples}
\end{figure}

We evaluate whether multi-relational alignment improves cross-modal retrieval over single-relation VLMs, and whether relation-aware learning benefits artwork classification across diverse
metadata types.

\subsection{Cross-modal retrieval}
\label{sec:cross_modal}
\Cref{tab:main_results_retrieval} reports retrieval results across the three datasets. \approach  achieves the best performance in image-to-text retrieval  across all datasets by a large margin, with image-to-text Recall@10  reaching 0.514 on \textsc{HertzianaDP}, 0.781 on \textsc{WikiArt+}, and 0.714 on \textsc{SemArt+}. In text-to-image retrieval, \approach achieves the best results on \textsc{HertzianaDP} and \textsc{SemArt+}, while remaining competitive on \textsc{WikiArt+}. These results confirm that modeling multiple relational subspaces yields richer representations that consistently benefit cross-modal alignment.  
The advantage is evident on  \textsc{HertzianaDP}, which is absent from CLIP's pretraining data, where \approach reaches an
I2T Recall@10 of 0.514, compared to 0.084 for CLIP-ft.  We note that, unlike CLIP, SigLIP, and MSC, which do not accept the relationship type, our textual prompt includes information about the relationship. This can be inferred by the model itself, as shown in \appcref{app:additional_results}, and does not require extra input information in practice. 
The I2T/T2I performance gap reflects a property of the data: images have multiple valid textual matches, giving I2T queries more correct targets, whereas a single text must retrieve a single image from among many similar alternatives. For instance, in \textsc{HertzianaDP} the test split has 6,874 images but 12,145 texts.

\Cref{fig:retrieval_split} provides a finer-grained view, breaking down T2I R@10 by relation type. \approach achieves strong performance across all relation types, whereas VLM baselines exhibit high variance, performing well on descriptive relations but degrading on more complex ones (\eg, context). 

\paragraph{Ablation} We ablate on the main components of \approach and assess their impact: (a) KL and InfoNCE losses, (b) graph-regularization without relation conditioning, and (c) simple multi-head projections per relation type. Results are in \cref{tab:ablations}. To test (a), we turn off first the KL and later the InfoNCE loss. We show that combining the two losses allows us to leverage the benefits of both (whereby the InfoNCE improves image-to-text retrieval, while the contextual-KL loss improves T2I). For (b), we remove the relationship-conditioned modulation and notice that it greatly degrades the model's performance. For (c), we show the need for the restriction maps by substituting them with simple per-relation linear heads and notice that the substitution penalizes the model.

\begin{table*}[t]
\centering
\caption{Ablations on \textsc{HertzianaDP}. The results are reported on \textsc{HertzianaDP} because it is the only dataset without potential data leakage from the base CLIP.}
\begin{tabular}{lcccc}
\toprule
Variant & T2I@5 & T2I@10 & I2T@5 & I2T@10 \\
\midrule
\approach & \textbf{0.099} & \textbf{0.158} & \underline{0.386} & \underline{0.514} \\
(a) only InfoNCE loss & 0.002 & 0.004 & \textbf{0.525} & \textbf{0.541} \\
(a) only KL loss & \underline{0.098} & \underline{0.156} & 0.385 & 0.490 \\
(b) no relation modulation & 0.002 & 0.003 & 0.001 & 0.002 \\
(c) simple multi-head projections & 0.021 & 0.040 & 0.140 & 0.170 \\
\bottomrule
\end{tabular}
\label{tab:ablations}
\end{table*}

\begin{table*}[t]
\centering
\caption{Classification accuracy across relation types on the selected datasets. \textbf{Bold} denotes the best model for a dataset, \underline{underlined} the second best.}
\label{tab:main_results_classification}
\resizebox{\linewidth}{!}{
\begin{tabular}{lcc|cccc|ccccc}
\toprule
 & \multicolumn{2}{c}{\textsc{HertzianaDP}} & \multicolumn{4}{c}{\textsc{SemArt+}} & \multicolumn{5}{c}{\textsc{WikiArt+}} \\
\cmidrule(lr){2-3}  \cmidrule(lr){4-7} \cmidrule(lr){8-12}
 Method & Artist & Period  & Artist & Date & Genre & Style & Artist & Genre & Style & Period & Material  \\
\midrule
ArtSAGENet & \textbf{0.808}  & \underline{0.791} & 0.121 & 0.563  & 0.520  & 0.374  & 0.430 & 0.023  & 0.563  & 0.218  & 0.089   \\
CLIP & 0.010  & 0.015 & 0.695  & 0.237  & 0.350  & 0.259 & 0.010  & 0.021  & 0.695  & 0.174  & 0.140  \\
CLIP-ft & 0.464 & 0.702 & 0.212 & 0.480 & 0.623 & 0.418 & 0.343 & 0.035 & 0.516 & \textbf{0.546} & 0.145  \\
ColPali & 0.014  & 0.082  & 0.128  & 0.109  & 0.166  & 0.102 & 0.014  & 0.008  & 0.128  & 0.084  & 0.005  \\
ColQwen & 0.020  & 0.028  & 0.645  & 0.036  & 0.025  & 0.014 & 0.020  & 0.010  & 0.645  & 0.062  & 0.011  \\
GraphCLIP & 0.588  & 0.719  & 0.074  & 0.381  & \textbf{0.738} & \underline{0.494} &  0.109  & 0.025  & 0.207  & \underline{0.522}  & 0.112  \\
MSC & 0.175  & 0.062  &  0.050  & 0.312  & 0.123  & 0.237 & 0.175  & 0.000  & 0.050  & 0.028  & 0.007   \\
RCML & 0.250 & 0.548 & 0.171 & 0.343 & 0.580 & 0.232 & 0.264 & 0.023 & 0.441 & 0.402 & 0.177 \\
SigLIP & 0.470  & 0.088  & \underline{0.702}  & 0.598  & 0.168  & 0.091 &  \textbf{0.470}  & 0.030  & \underline{0.702}  & 0.257  & \underline{0.283}   \\
SigLIP-ft & \underline{0.650} & 0.776 & 0.620 & \underline{0.611} & 0.585 & 0.459 & 0.460 & \underline{0.060} & 0.644 & 0.339 & \textbf{0.291} \\
\approach & 0.472  & \textbf{0.794}  & \textbf{0.808} & \textbf{0.634} &\underline{ 0.628} & \textbf{0.637} & \underline{0.462}  & \textbf{0.043}  & \textbf{0.790}  & 0.485  & 0.226 \\
\bottomrule
\end{tabular}
}
\end{table*}

\subsection{Relation-aware classification}

\Cref{tab:main_results_classification} reports classification accuracy across relation types. \approach achieves the best or second-best result on almost every task. Crucially, no single baseline matches this consistency: when \approach ranks second, the top-performing method varies across
tasks, meaning that no alternative baseline can be selected as a uniformly better choice. \approach consistently outperforms all VLM baselines on relationally complex attributes such as Style and Period, where single-embedding methods struggle. Compared to the graph-based ArtSAGENet and MSC, \approach achieves superior performance on the majority of tasks while operating inductively, without requiring graph connectivity at test time. 

\subsection{Qualitative analysis}
\Cref{fig:qualitative_examples} illustrates how \approach modulates
retrieval depending on the relation on \textsc{WikiArt+}. For Matisse's
\emph{Sheaf}, the textual queries retrieve different aspects related to the painting and artist: its gestural nature of abstract expressionism is retrieved through Rosenberg's quote; a title query surfaces Cubist compositions; the artist query returns Fauvist-adjacent works; the date yields contemporaneous 50's pieces. Image-to-Image retrieval exhibits analogous behavior, confirming that each subspace captures distinct art-historical dimensions.

\begin{figure}[t]
    \centering
    \includegraphics[width=\linewidth]{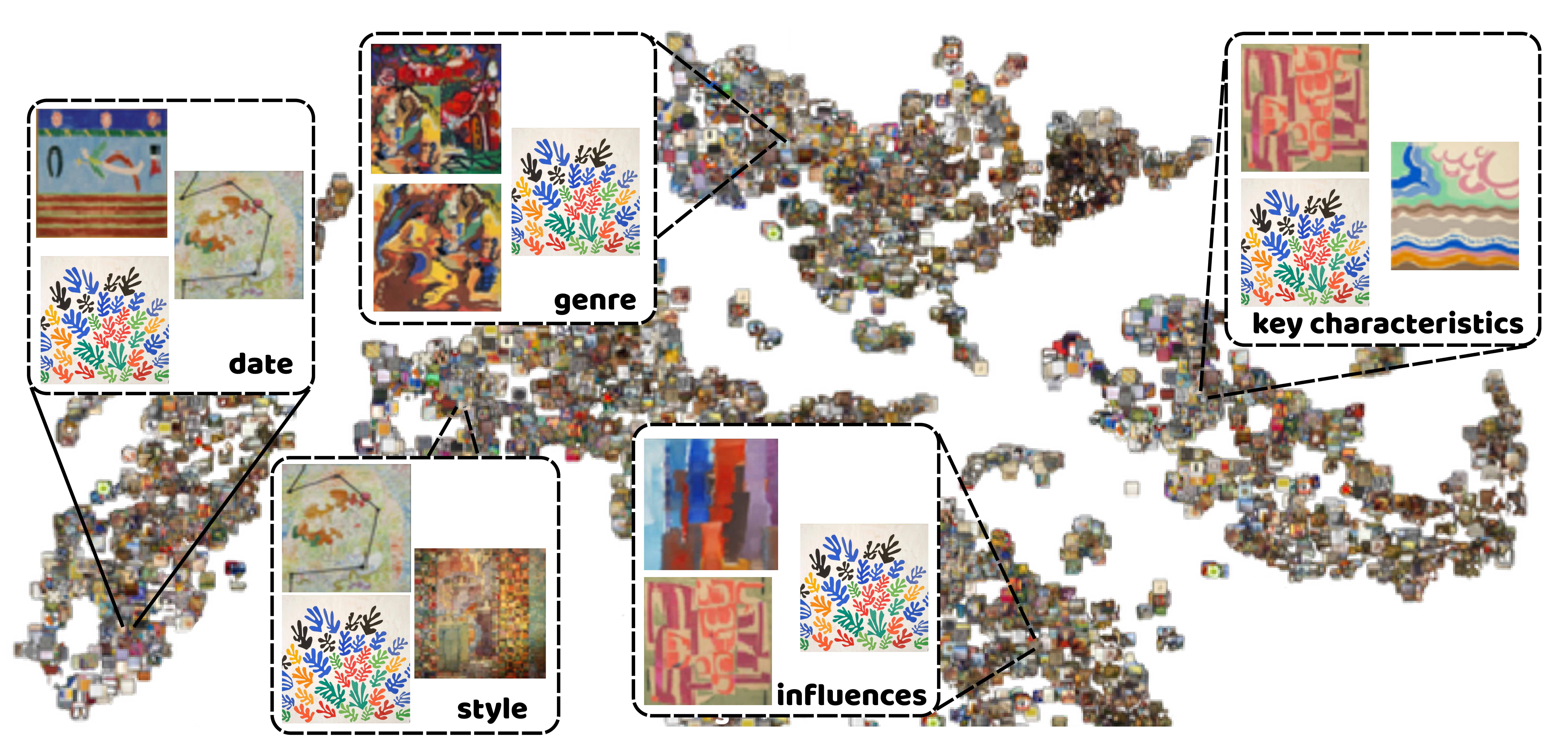}
    \caption{\textbf{Visualization of relation-aware embedding space learned by \approach on the \textsc{WikiArt+} dataset}. The plot shows a 2D t-SNE projection of the learned multimodal embeddings, with points representing image embeddings. The image shows how \approach organizes artworks into semantically meaningful clusters based on different relations. The inset boxes highlight specific relational neighborhoods around Matisse's ``Sheaf'': artworks clustered by temporal information show examples of art from the 50's, stylistic neighbors are at the boundary between abstract expressionism and Art Nouveau, the genre highlights a tension between figurative and abstract art, while influences are primarily from abstract expressionism, with key characteristics mirroring the same tension between recognizable figures and color planes.}
    \label{fig:csn}
\end{figure}

\Cref{fig:csn} provides geometric evidence via a t-SNE projection~\cite{maaten2008visualizing}
on \textsc{WikiArt+}. Around Matisse's \emph{Sheaf}, different relational
lenses yield different neighborhoods: 50's works for date,
Abstract Expressionist/Art Nouveau neighborhood of style, artworks at the boundary between figurative and abstract for genre, and partly figurative flat-color-plane compositions for key characteristics. These neighborhoods are neither identical nor redundant, confirming that \approach maintains complementary subspaces, which substantiates the results in
\cref{tab:main_results_retrieval,tab:main_results_classification}.


\section{Conclusion}
We presented \approach, a sheaf-inspired framework learning relation-aware multimodal representations for art understanding. Through aspect-conditioned embeddings and a graph-informed contrastive loss, it captures multi-relational structure while remaining fully inductive at inference.
Together with three new benchmarks, \textsc{WikiArt+}, \textsc{SemArt+} and \textsc{HertzianaDP}, experiments across datasets demonstrate improvements over VLM and graph-based baselines.

Beyond art understanding, \approach is domain-agnostic and suits settings such as medical imaging and cultural heritage, where images and texts share heterogeneous semantic relations.

\section*{Acknowledgements}
The authors acknowledge Pietro Liuzzo and the Photographic Collection team of the Bibliotheca Hertziana for the support in providing the \textsc{HertzianaDP} dataset. Antonio Purificato and Fabrizio Silvestri acknowledge project EVOLVE: A Fluid Framework for Dynamic Agentic AI Systems, Progetto Ateneo Sapienza. Noa Garcia acknowledges JSPS KAKENHI No. 23H00497 and No. 22K12091.

\newpage
\bibliographystyle{splncs04}
\bibliography{biblio}
\newpage

\section*{Supplementary Material}

\appendix
\section{Experimental details}
\label{app:exp_setup}

\begin{table}[ht]
\caption{Number of images, texts, nodes, and edges in the datasets for train, validation, and test, respectively.}
    \centering
    \resizebox{\linewidth}{!}{
    \begin{tabular}{ccccc|cccc|cccc}
    \toprule
    & \multicolumn{4}{c}{Train} & \multicolumn{4}{c}{Val} & \multicolumn{4}{c}{Test} \\
    \cmidrule(lr){2-5}  \cmidrule(lr){6-9} \cmidrule(lr){10-13}
     Dataset & Images & Texts & Nodes & Edges & Images & Texts & Nodes & Edges & Images & Texts & Nodes & Edges \\
    \midrule
     \textsc{HertzianaDP} & 32,073 & 54,917 & 86,990 & 133,179 & 6,873 & 12,385 & 19,258& 28,637 & 6,874 & 12,145& 19,019 & 28,379\\
     \textsc{SemArt+} & 19,243 & 40,259 & 59,502 & 115,818  & 14,458 & 16,855 & 31,313 & 25,701 & 1,069 & 5,169 & 6,238 & 9,911 \\
     \textsc{WikiArt+} & 15,419 & 17,509 & 32,928 & 215,855 & 3,304 & 6,065 & 9369 & 46,222 & 3,305 & 6,029 & 9,334 & 45,973 \\
    \bottomrule
    \end{tabular}
    }
    \label{tab:datasets_stats}
\end{table}

\subsection{Experimental setup}
Our experiments are performed on a workstation equipped with an Intel Core i9-10940X (14-core CPU running at 3.3 GHz), 256 GB of RAM, and a single Nvidia RTX A6000 with 48 GB of VRAM.

\subsection{Datasets}

We evaluate our approach on three datasets, whose statistics are summarized in \cref{tab:datasets_stats}:
\begin{itemize}
    \item \textsc{HertzianaDP}: it is the largest corpus by total nodes, comprising over 86,000 training nodes derived from 32,073 images and 54,917 textual descriptions. Each image is associated with, on average, multiple captions, resulting in a densely connected graph with 133,179 training edges.
    \item \textsc{SemArt+}: it contains 19,243 training images paired with 40,259 textual annotations, yielding 115,818 edges in the
    training split. 
    \item \textsc{WikiArt+}: it is the most edge-dense dataset relative to its number of nodes: despite having only 15,419 training images and 17,509 texts, the training graph contains 215,855 edges, indicating a high degree of interconnection among artworks and their associated metadata.
\end{itemize}
Across all three datasets, the data is split into training, validation, and test partitions. For \textsc{HertzianaDP} and \textsc{WikiArt+}, we adopt a 70/15/15 train/validation/test split in terms of images.
For \textsc{SemArt+}, we follow the original split provided
by the authors~\cite{garcia2018read}.
The multimodal graph structure naturally arises from the heterogeneous relationships linking visual and textual nodes, where each edge represents a semantic association (\eg, authorship, genre, or temporal period) between an artwork and its descriptive attributes.

\subsection{Pronoun resolution for \textsc{SemArt+}}
\label{sec:pronounres}

The \textsc{SemArt+} dataset augments \textsc{SemArt} \cite{garcia2018read} with the per-sentence aspect annotations introduced by \textsc{Explain Me the Painting} \cite{bai2021explain}. It extracts sentences corresponding to different aspects of a painting (\eg, ``context'') from a full textual description.
These sentences frequently contain pronouns referring to entities introduced earlier in the description. However, in our task, each sentence is considered in isolation, which may introduce ambiguity when the subject is not explicitly repeated. For example:

\begin{tcolorbox}[colback=gray!10, colframe=gray!50,
width=\textwidth]
\small{
{\ttfamily {\texttt{<content>} \textit{The painting depicts a ``forest floor'' still-life of plants and flowers by a wall with an urn; beyond are elegant figures outside a classical mansion. Begeyn is a versatile painter who has produced numerous landscapes.\texttt{</content>} \texttt{<context>}Occasionally, he devotes almost all of his subject to the ``forest floor'' in a manner reminiscent of Otto Marseus van Schrieck} \texttt{</context>}
} }}
\end{tcolorbox}

In this example, directly extracting the \texttt{<context>} sentence makes it difficult to interpret the meaning of the text, since the subject is only referred to through the pronoun ``he''. To address this issue, we perform pronoun resolution on the \textsc{SemArt+} dataset using GPT-5-nano\footnote{\url{https://developers.openai.com/api/docs/models/gpt-5-nano}}. 

Before processing the descriptions, we introduced explicit labels (\eg, \texttt{<content>} and \texttt{</content>}) around the sentences corresponding to specific interpretations of the painting. The descriptions were then processed while preserving these labels. 
The resulting description after pronoun resolution would be:

\begin{tcolorbox}[colback=gray!10, colframe=gray!50,
width=\textwidth]
\small{
{\ttfamily {\texttt{<content>} \textit{The painting depicts a ``forest floor'' still-life of plants and flowers by a wall with an urn; beyond are elegant figures outside a classical mansion. Begeyn is a versatile painter who has produced numerous landscapes.\texttt{</content>} \texttt{<context>}Occasionally, \textbf{Begeyn} devotes almost all of\textbf{ Begeyn's} subject to the ``forest floor'' in a manner reminiscent of Otto Marseus van Schrieck} \texttt{</context>}
} }}
\end{tcolorbox}

The system and user prompts used for this task are:

\begin{tcolorbox}[colback=gray!10, colframe=gray!50,
width=\textwidth]
\small{
{\ttfamily {
prompt\_system = f"""
You are a careful text editor specializing in art history writing.
Your task is to perform *reference resolution* on the text below. Replace pronouns
(e.g., "he," "she," "it," "they," "his," "their," etc.) and **generic referential noun phrases**
(e.g., "the artist," "the painter," "the work," "the century," "the movement") with their most specific explicit referents when doing so is clear and natural.

Critical accuracy rules:
\begin{itemize}
    \item Do **not** remove, shorten, paraphrase, or summarize the text.
    \item Do **not** add or invent any information.
    \item  Keep **all original wording and punctuation** except where replacing a reference.
    \item **Preserve all existing labels** such as `<content>...</content>`, `<context>...</context>`, `<form>...</form>` exactly in their original positions.
    \item You may modify the text inside the labels as needed to resolve references, but **do not move, remove, or split any labels**.
    \item Do **not** add new labels.
    \item Only resolve references when the intended referent is **certain**.
    \item If uncertain, **leave the original text unchanged**.
    \item **Resolve across sentences and across the paragraph** - a referring term may relate to something introduced earlier.
    \item **Do not resolve a reference if the specific noun already appears in the same sentence**, unless this improves clarity without causing unnatural repetition.
    \item Avoid unnatural repetition (e.g., avoid things like ``The artist admired the artist's paintings.'')
    \item Prefer clarity and specificity, while keeping the style natural and faithful to the original.
    \item Output **only the rewritten text**, preserving all labels exactly, without explanation.
    \end{itemize}
"""

\begin{verbatim}
def make_prompt_user(ref_test_text):
  prompt_user = f"Text to rewrite: {ref_test_text}"
  return prompt_user
\end{verbatim}
   
} }}
\end{tcolorbox}

\subsection{Computational cost}

\begin{table}[ht]
    \caption{Computational cost (in TFLOPS) for each model.}
    \centering
    \begin{tabular}{cc}
    \toprule
    Model & TFLOPS \\
    \midrule
    ArtSAGENet & 1.89 \\
    CLIP & 0.01 \\
    ColPali &  4.93 \\
    ColQwen & 2.02 \\
    MSC & 1.05 \\
    SigLIP  & 0.22\\
    \approach & 1.37 \\
    \bottomrule
    \end{tabular}
    \label{tab:flops}
\end{table}

\cref{tab:flops} reports the computational cost of each model in terms of tera floating-point operations (TFLOPS).
CLIP is by far the most lightweight model, requiring only 0.01 TFLOPS, followed by SigLIP (0.22 TFLOPS), which remains relatively efficient despite its larger vision backbone. MSC and \approach occupy a similar mid-range tier at 1.05 and 1.37 TFLOPS, respectively, while ArtSAGENet requires 1.89 TFLOPS due to its graph neural network overhead on top of the visual encoder. ColQwen and ColPali are the most expensive models, the latter being nearly 500$\times$ more costly than CLIP, reflecting the heavier computational burden of
late-interaction multi-vector retrieval architectures.
\approach achieves a favorable trade-off between performance and computational cost, operating at roughly one-fourth of the cost of ColPali while remaining competitive in retrieval accuracy.

It is worth noting that a significant portion of the computational cost of graph-based methods such as ArtSAGENet and \approach is attributable to the message-passing operations along the edge index, whose complexity scales with the number of edges in the graph. In contrast, the cost of pairwise models such as CLIP and SigLIP depends solely on the number of samples, regardless of graph connectivity. 

\section{Additional results}
\label{app:additional_results}

Results in terms of Precision and NDCG, reported in~\cref{tab:precision_results,tab:ndcg_results}, are consistent with the trends observed for Recall in the main paper. In particular, \approach consistently achieves the best performance in the I2T retrieval task across all datasets and cut-off values, with a substantial margin over the competing methods. The improvements are especially pronounced on \textsc{WikiArt+} and \textsc{SemArt+}, where \approach markedly outperforms all baselines in both P@K and NDCG@K. Similar gains are observed on \textsc{HertzianaDP} for I2T retrieval. For T2I retrieval, the performance differences among models are generally smaller; nevertheless, \approach remains competitive and achieves the best results on several settings, particularly on \textsc{WikiArt+} and \textsc{SemArt+}. Overall, these results further confirm the effectiveness of \approach and align with the recall-based analysis presented in the main paper.

\begin{table*}[t]
\caption{Retrieval results in terms of Precision (P) in text-to-image (T2I) and image-to-text (I2T) across the selected datasets ($K=\{5,10\}$). \textbf{Bold} denotes the best model for a dataset, \underline{underlined} the second best.}\centering
\resizebox{\linewidth}{!}{
\begin{tabular}{lcccc|cccc|cccc}
\toprule
& \multicolumn{4}{c}{\textsc{HertzianaDP}} & \multicolumn{4}{c}{\textsc{SemArt+}} & \multicolumn{4}{c}{\textsc{WikiArt+}} \\
\cmidrule(lr){2-5}  \cmidrule(lr){6-9} \cmidrule(lr){10-13}
 Method & T2I P@5 & T2I P@10 & I2T P@5 & I2T P@10 & T2I P@5 & T2I P@10 & I2T P@5 & I2T P@10 & T2I P@5 & T2I P@10 & I2T P@5 & I2T P@10 \\
\midrule
ArtSAGENet & 0.000 & 0.000 & 0.000 & 0.000 & 0.002 & 0.002 & 0.000 & 0.000 & 0.001 & 0.001 & 0.001 & 0.001 \\
CLIP & 0.003 & 0.002 & 0.004 & 0.003 & 0.030 & 0.018 & 0.021 & 0.015 & 0.013 & 0.008 & 0.011 & 0.008  \\
CLIP-ft & 0.005 & 0.003 & 0.004 & 0.003 & 0.032 & 0.021 & 0.023 & 0.017 & 0.142 & 0.017 & 0.032 & 0.021 \\
ColPali & 0.001 & 0.001 & 0.001 & 0.001 & 0.010 & 0.007 & 0.005 & 0.004 & 0.003 & 0.002 & 0.002 & 0.001 \\
ColQwen & 0.001 & 0.000 & 0.000 & 0.000 & 0.016 & 0.009 & 0.007 & 0.005 & 0.003 & 0.002 & 0.001 & 0.001 \\
MSC & 0.000 & 0.000 & 0.000 & 0.000 & 0.000 & 0.000 & 0.011 & 0.009 & 0.000 & 0.000 & 0.009 & 0.005 \\
SigLIP & \underline{0.006} & \underline{0.004} & 0.005 & 0.003 & 0.034 & \underline{0.021} & 0.024 & 0.017 & 0.023 & 0.015 & 0.016 & 0.011  \\
SigLIP-ft & \textbf{0.007} & \textbf{0.005} & \underline{0.006} & \underline{0.007} & \textbf{0.039} & 0.021 & \underline{0.059} & \underline{0.039} & \underline{0.091} & \underline{0.087} & \underline{0.093} & \textbf{0.082} \\
\approach & 0.001 & 0.001 & \textbf{0.093} & \textbf{0.049} & \underline{0.037} & \textbf{0.027} & \textbf{0.122} & \textbf{0.068} & \textbf{0.128} & \textbf{0.110} & \textbf{0.132} & \underline{0.073} \\
\bottomrule
\end{tabular}
}
\label{tab:precision_results}
\end{table*}

\begin{table*}[t]
\caption{Retrieval results in terms of NDCG (N) in text-to-image (T2I) and image-to-text (I2T) across the selected datasets ($K=\{5,10\}$). \textbf{Bold} denotes the best model for a dataset, \underline{underlined} the second best.}
\centering
\resizebox{\linewidth}{!}{
\begin{tabular}{lcccc|cccc|cccc}
\toprule
& \multicolumn{4}{c}{\textsc{HertzianaDP}} & \multicolumn{4}{c}{\textsc{WikiArt+}} & \multicolumn{4}{c}{\textsc{SemArt+}} \\
\cmidrule(lr){2-5}  \cmidrule(lr){6-9} \cmidrule(lr){10-13}
 Method & T2I N@5 & T2I N@10 & I2T N@5 & I2T N@10 & T2I N@5 & T2I N@10 & I2T N@5 & I2T N@10 & T2I N@5 & T2I N@10 & I2T N@5 & I2T N@10 \\
\midrule
ArtSAGENet & 0.001 & 0.001 & 0.000 & 0.000 & 0.001 & 0.002 & 0.001 & 0.001 & 0.002 & 0.003 & 0.000 & 0.000 \\
CLIP & 0.008 & 0.010 & 0.013 & 0.015 & 0.102 & 0.072 & 0.051 & 0.061 & 0.016 & 0.020 & 0.039 & 0.047 \\
CLIP-ft & 0.010 & 0.011 & 0.022 & 0.031 & 0.106 & 0.112 & 0.124 & 0.188 & 0.033 & 0.058 & \underline{0.153} & \underline{0.212} \\
ColPali & 0.003 & 0.003 & 0.003 & 0.006 & 0.036 & 0.025 & 0.014 & 0.016 & 0.004 & 0.007 & 0.006 & 0.008 \\
ColQwen & 0.002 & 0.002 & 0.001 & 0.002 & 0.055 & 0.037 & 0.020 & 0.023 & 0.004 & 0.006 & 0.005 & 0.006 \\
MSC & 0.000 & 0.000 & 0.000 & 0.000 & 0.001 & 0.001 & 0.039 & 0.049 & 0.000 & 0.000 & 0.035 & 0.038 \\
SigLIP & \underline{0.016} & \underline{0.018} & 0.017 & 0.020 & 0.109 & 0.086 & 0.065 & 0.077 & 0.032 & 0.040 & 0.058 & 0.067  \\
SigLIP-ft & \textbf{0.018} & \textbf{0.021} & \underline{0.107} & \underline{0.152} & \textbf{0.122} & \underline{0.124} & \underline{0.253} & \underline{0.239} & \underline{0.054} & \underline{0.089} & 0.121 & 0.189 \\
\approach & 0.001 & 0.001 & \textbf{0.417} & \textbf{0.423} & \underline{0.111} & \textbf{0.127} & \textbf{0.513} & \textbf{0.532} & \textbf{0.162} & \textbf{0.172} & \textbf{0.553} & \textbf{0.575} \\
\bottomrule
\end{tabular}
}
\label{tab:ndcg_results}
\end{table*}

\subsection{Relationship prediction and comparability}

\begin{table*}[h]
\centering
\caption{Classification results on the relationship prediction. Accuracy refers to the test accuracy across classes based on the text.}
\begin{tabular}{lc}
\toprule
Dataset & Accuracy \\
\midrule
\textsc{HertzianaDP} & 1.00 \\
\textsc{SemArt+} & 0.86  \\
\textsc{WikiArt+} & 0.99  \\
\bottomrule
\end{tabular}
\label{tab:link_pred_accuracy}
\end{table*}

As noted in~\cref{sec:cross_modal}, cross-modal retrieval across the entire test dataset leverages relationship information that not all baselines can accommodate. To ensure comparability, in this section, we show that (i) even including relationship information in the text of CLIP and SigLIP does not improve their performance, (ii) the relationship can easily be inferred using a simple classifier on CLIP embeddings, and the use of predicted relationship type does not significantly decrease the performance of our model. 

In \cref{tab:link_info}, we add relationship information in the texts passed to CLIP and SigLIP, which previously did not receive this information. We compose the new text as \texttt{``\{relationship\}: \{text\}''}. The inclusion of the relationship type does not improve retrieval. 

\begin{table}[ht]
\caption{Additional results to Table 1. Retrieval results in terms of Recall@5 (R@5). \textbf{Bold} denotes the best model for a dataset, \underline{underlined} the second best. \textbf{L-} indicates relationship (link type) information has been added to the text for better comparability. 
}
\centering

\begin{tabular}{llccc}
\toprule
Model & Variant & \textsc{HertzianaDP} & \textsc{SemArt+} & \textsc{WikiArt+} \\
\midrule
\multirow{2}{*}{\approach}
 & T2I & \textbf{0.099} & \textbf{0.376} & \underline{0.114} \\
 & I2T & \textbf{0.386} & \textbf{0.645} & \textbf{0.703} \\
\midrule
\multirow{6}{*}{CLIP}
 & T2I    & 0.041 & 0.305 & 0.131 \\
 & \textbf{L}-T2I  & 0.036 & 0.294 & 0.100 \\
 & I2T    & 0.013 & 0.078 & 0.042 \\
 & \textbf{L}-I2T  & 0.013 & 0.079 & 0.034 \\
\midrule
\multirow{6}{*}{SigLIP}
 & T2I    & \underline{0.061} & \underline{0.334} & \textbf{0.239} \\
 & \textbf{L}-T2I  & 0.051 & 0.310 & 0.169 \\
 & I2T    & 0.016 & 0.095 & 0.067 \\
 & \textbf{L}-I2T  & 0.016 & 0.078 & 0.053 \\
\bottomrule
\end{tabular}
\label{tab:link_info}
\end{table}

Furthermore, in \cref{tab:link_pred_accuracy}, we show that the relationship type can easily be inferred from the text itself. We train a simple two-layer classifier on top of the CLIP embeddings of the texts in the training set. The target is the relationship that the text entertains with the images. The test accuracy shows that this can be easily inferred at test time without requiring additional information. 

To further strengthen this claim, in \cref{tab:predicted}, we show that using predicted links rather than the original ones does not significantly degrade our model's performance, confirming that we do not require additional information at test time and ensuring a fully inductive setting.

\begin{table*}[t]
\centering
\caption{Comparison between \approach using the original relationship type information and \approach using the predicted information from the model in \cref{tab:link_pred_accuracy}. The performance is evaluated on the original dataset. We do not observe significant differences between the modalities.}
\label{tab:predicted}
\resizebox{\linewidth}{!}{
\begin{tabular}{lcccc|cccc|cccc}
\toprule
& \multicolumn{4}{c}{\textsc{HertzianaDP}} & \multicolumn{4}{c}{\textsc{SemArt+}} & \multicolumn{4}{c}{\textsc{WikiArt+}} \\
\cmidrule(lr){2-5}  \cmidrule(lr){6-9} \cmidrule(lr){10-13}
 Method & T2I R@5 & T2I R@10 & I2T R@5 & I2T R@10 & T2I R@5 & T2I R@10 & I2T R@5 & I2T R@10 & T2I R@5 & T2I R@10 & I2T R@5 & I2T R@10 \\
\midrule
L-PRED. & 0.105 & 0.165 & 0.412 & 0.578 & 0.313 & 0.414 & 0.636 & 0.703 & 0.146 & 0.224 & 0.704 & 0.776  \\
L-ORIG. & 0.099 & 0.158 & 0.386 & 0.514 & 0.376 & 0.489 & 0.645 & 0.714 & 0.144 & 0.217 & 0.703 & 0.781 \\
\bottomrule
\end{tabular}
}
\end{table*}

\begin{figure*}[h]
\begin{subfigure}{0.3\textwidth}
    \centering
    \includegraphics[width=\textwidth]{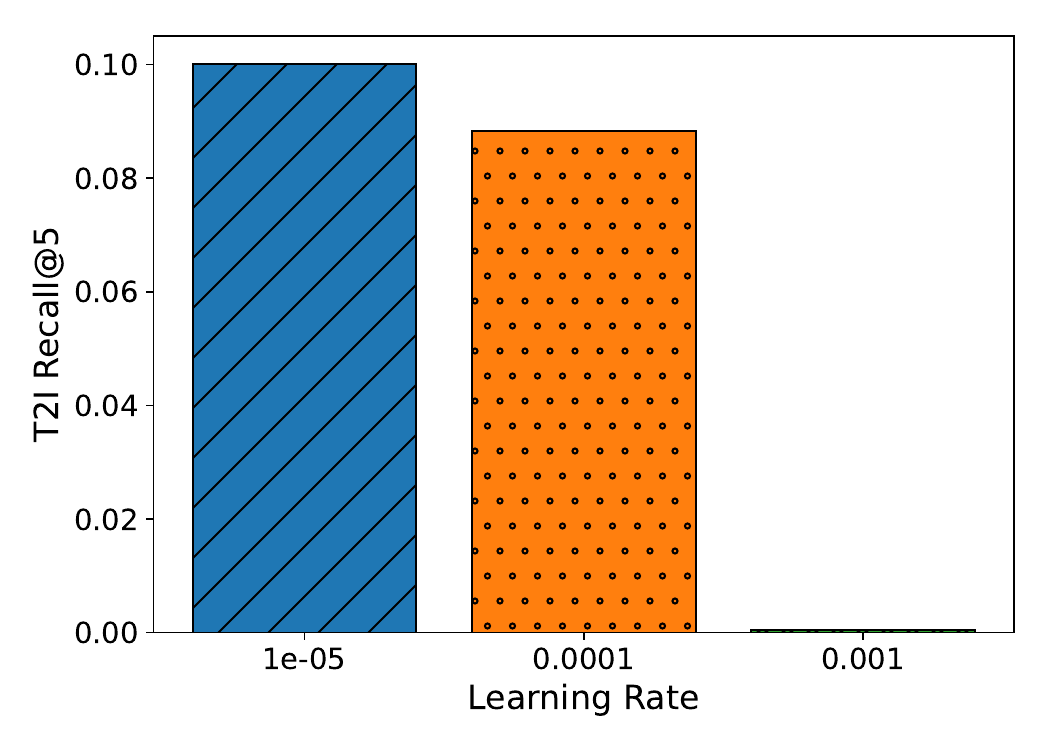}
\end{subfigure} 
\begin{subfigure}{0.3\textwidth}
    \centering
    \includegraphics[width=\textwidth]{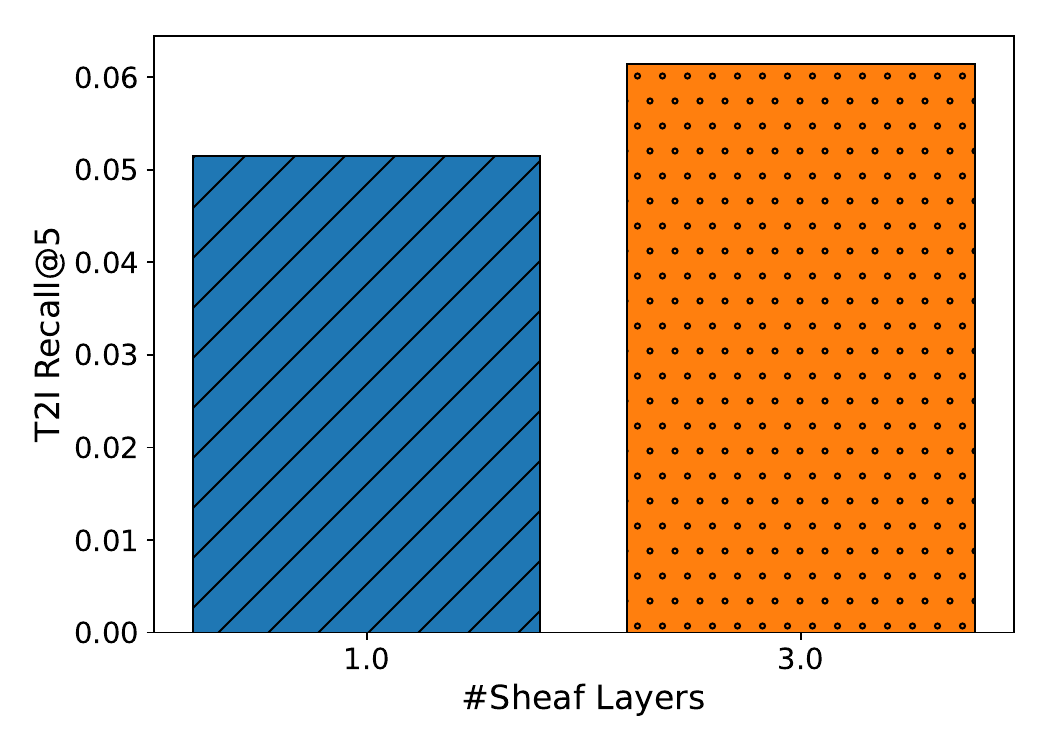}
\end{subfigure} 
\begin{subfigure}{0.3\textwidth}
    \centering
    \includegraphics[width=\textwidth]{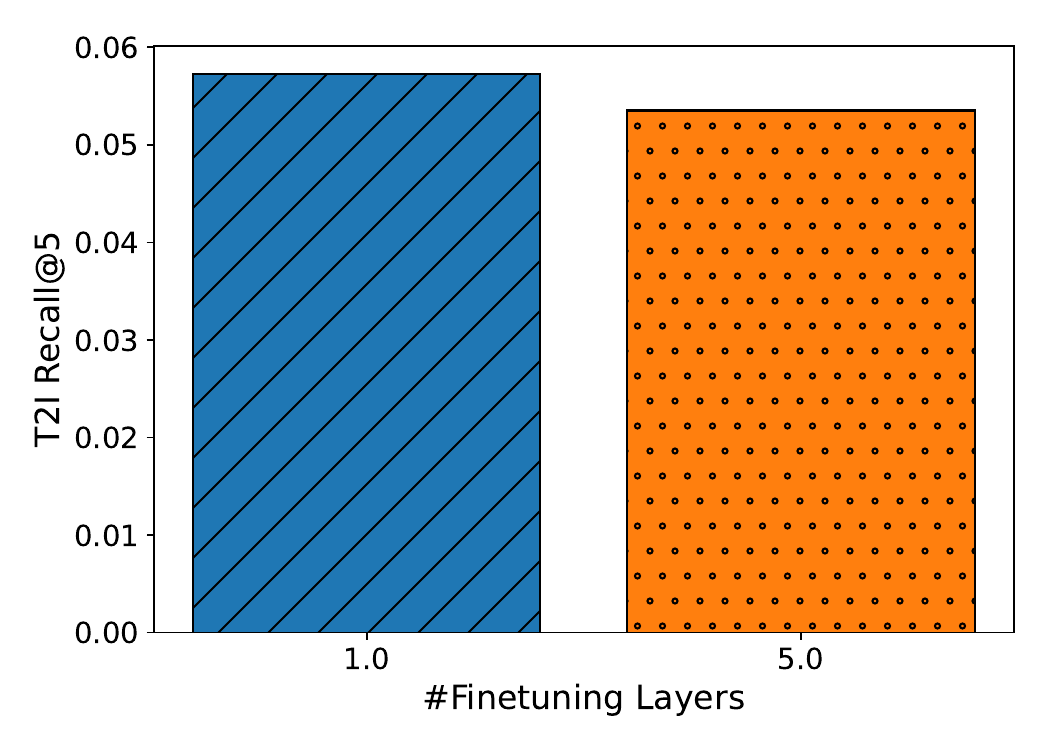}
\end{subfigure} \\
\begin{subfigure}{0.3\textwidth}
    \centering
    \includegraphics[width=\textwidth]{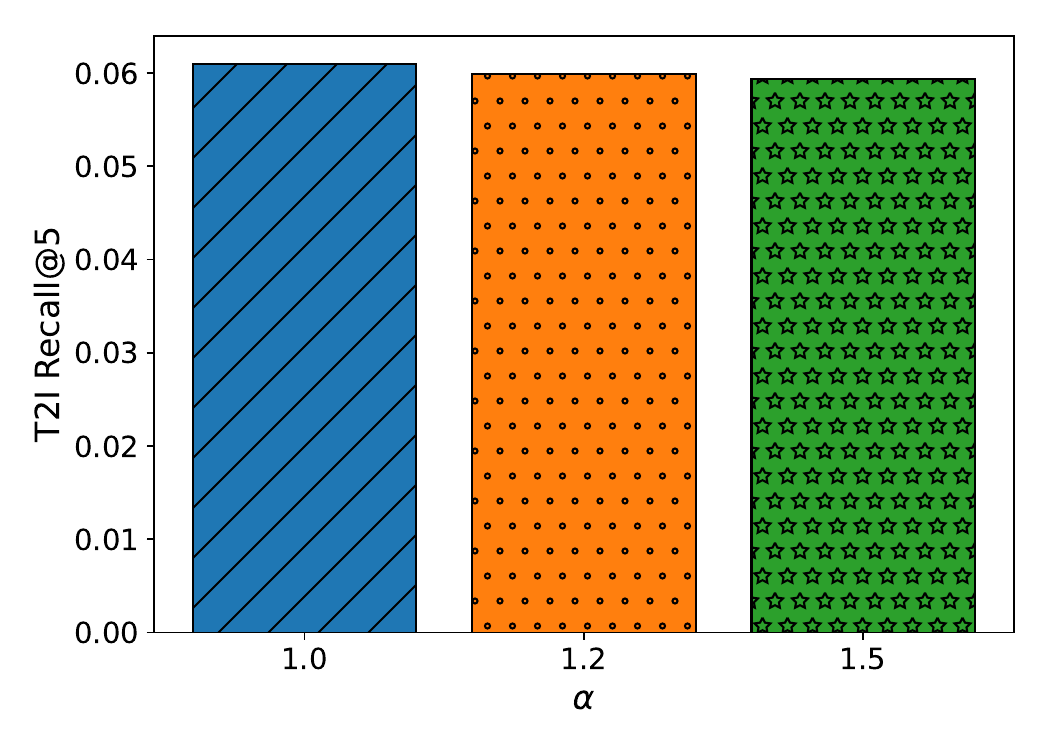}
\end{subfigure} 
\begin{subfigure}{0.3\textwidth}
    \centering
    \includegraphics[width=\textwidth]{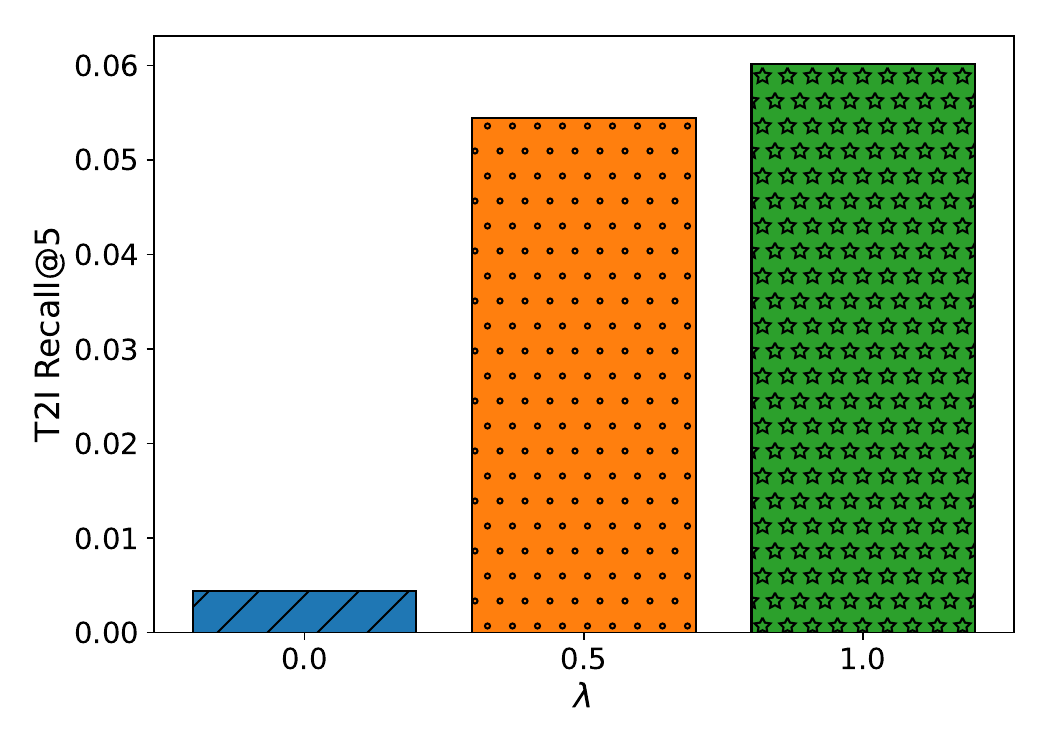}
\end{subfigure} 
\begin{subfigure}{0.3\textwidth}
    \centering
    \includegraphics[width=\textwidth]{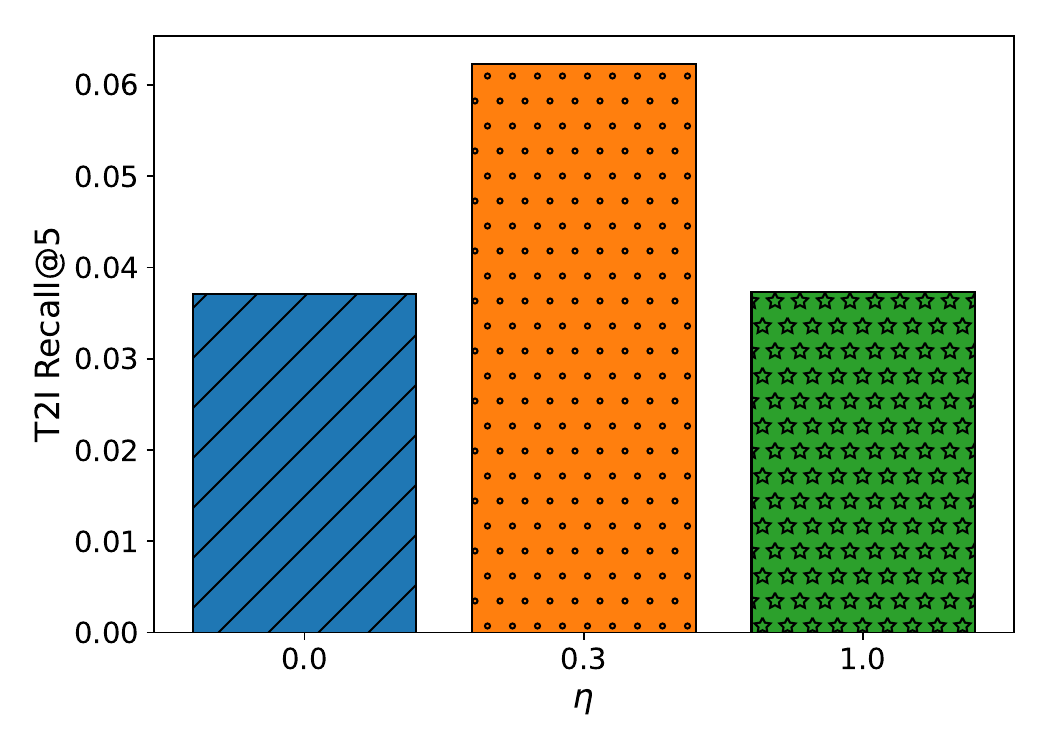}
\end{subfigure}
\caption{Hyperparameter sensitivity analysis of \approach on \textsc{HertzianaDP} in terms of T2I Recall@5.}
\label{fig:sensitivity}
\end{figure*}

\subsection{Hyperparameters optimization}

\Cref{fig:sensitivity} reports the sensitivity of \approach to its key hyperparameters on \textsc{HertzianaDP}, measured as  T2I Recall@5. Each subplot varies one hyperparameter while keeping the others fixed. Overall, the model proves robust to most hyperparameters, with performance remaining stable across a wide range of values for the concentration parameter  $\alpha$, the number of sheaf layers, and the number of fine-tuning layers. The most sensitive parameters are the learning rate, where larger values cause a sharp performance drop, the KL loss weight $\lambda$, whose removal confirms the ablation findings, and the loss weight $\eta$, 
which exhibits a clear optimum at intermediate values. 

This analysis guides the selection of the final hyperparameter configuration used throughout all experiments reported in this paper.

\end{document}